\documentclass[conference]{IEEEtran}
\usepackage{times}

\usepackage[numbers]{natbib}
\usepackage{multicol}
\usepackage{xcolor}
\usepackage[colorlinks,linkcolor=blue]{hyperref}
\usepackage{graphicx}
\usepackage{amsmath}
\usepackage{amsfonts}
\usepackage{algpseudocode}
\usepackage{algorithm}
\let\labelindent\relax
\usepackage{enumitem}
\usepackage{tikz}
\usepackage{multirow}
\usepackage{booktabs}
\usepackage{tcolorbox}

\begin{document}


\title{
    FUNCTO: Function-Centric One-Shot \\ Imitation Learning for Tool Manipulation
}


\author{
\authorblockN{Chao Tang$^{1 2}$, Anxing Xiao$^2$, Yuhong Deng$^2$, Tianrun Hu$^3$, Wenlong Dong$^1$, \\Hanbo Zhang$^2$, David Hsu$^{23}$, and Hong Zhang$^1$}
\authorblockA{$^1$Southern University of Science and Technology   $^2$School of Computing, National University of Singapore}
\authorblockA{$^3$Smart Systems Institute, National University of Singapore}
}



%

\maketitle

\begin{abstract}
Learning tool use from a single human demonstration video offers a highly intuitive and efficient approach to robot teaching. While humans can effortlessly generalize a demonstrated tool manipulation skill to diverse tools that support the same function (e.g., pouring with a mug versus a teapot), current one-shot imitation learning (OSIL) methods struggle to achieve this. A key challenge lies in establishing functional correspondences between demonstration and test tools, considering significant geometric variations among tools with the same function (i.e., intra-function variations). To address this challenge, we propose FUNCTO (\underline{Func}tion-Centric OSIL for \underline{To}ol Manipulation), an OSIL method that establishes function-centric correspondences with a 3D functional keypoint representation, enabling robots to generalize tool manipulation skills from a single human demonstration video to novel tools with the same function despite significant intra-function variations. With this formulation, we factorize FUNCTO into three stages: (1) functional keypoint extraction, (2) function-centric correspondence establishment, and (3) functional keypoint-based action planning. We evaluate FUNCTO against exiting modular OSIL methods and end-to-end behavioral cloning methods through real-robot experiments on diverse tool manipulation tasks. The results demonstrate the superiority of FUNCTO when generalizing to novel tools with intra-function geometric variations. More details are available at \href{https://sites.google.com/view/functo}{https://sites.google.com/view/functo}.
\end{abstract}


\IEEEpeerreviewmaketitle

\section{Introduction}


The ability to use tools has long been recognized as a hallmark of human intelligence \cite{washburn1960tools}. 
Endowing robots with the same capability holds the promise of unlocking a wide range of downstream tasks and applications \cite{chi2023diffusion, finn2017one, vitiello2023one}. As a step towards this goal, we tackle the problem of one-shot imitation learning (OSIL) for tool manipulation, which involves teaching robots a tool manipulation skill with a single human demonstration video. The objective is to develop an OSIL method capable of \textit{generalizing the demonstrated tool manipulation skill to novel tools with the same function.} Here, ``same function" refers to the robot imitating the demonstrated tool manipulation behavior to accomplish functionally equivalent tasks.





\begin{figure}[th]
  \centering
  \vspace*{-0.1in}
  \begin{tikzpicture}[inner sep = 0pt, outer sep = 0pt]
    \node[anchor=south west] (fnC) at (0in,0in)
      {\includegraphics[height=5.2in,clip=true,trim=0in 0in 0.1in 0in]{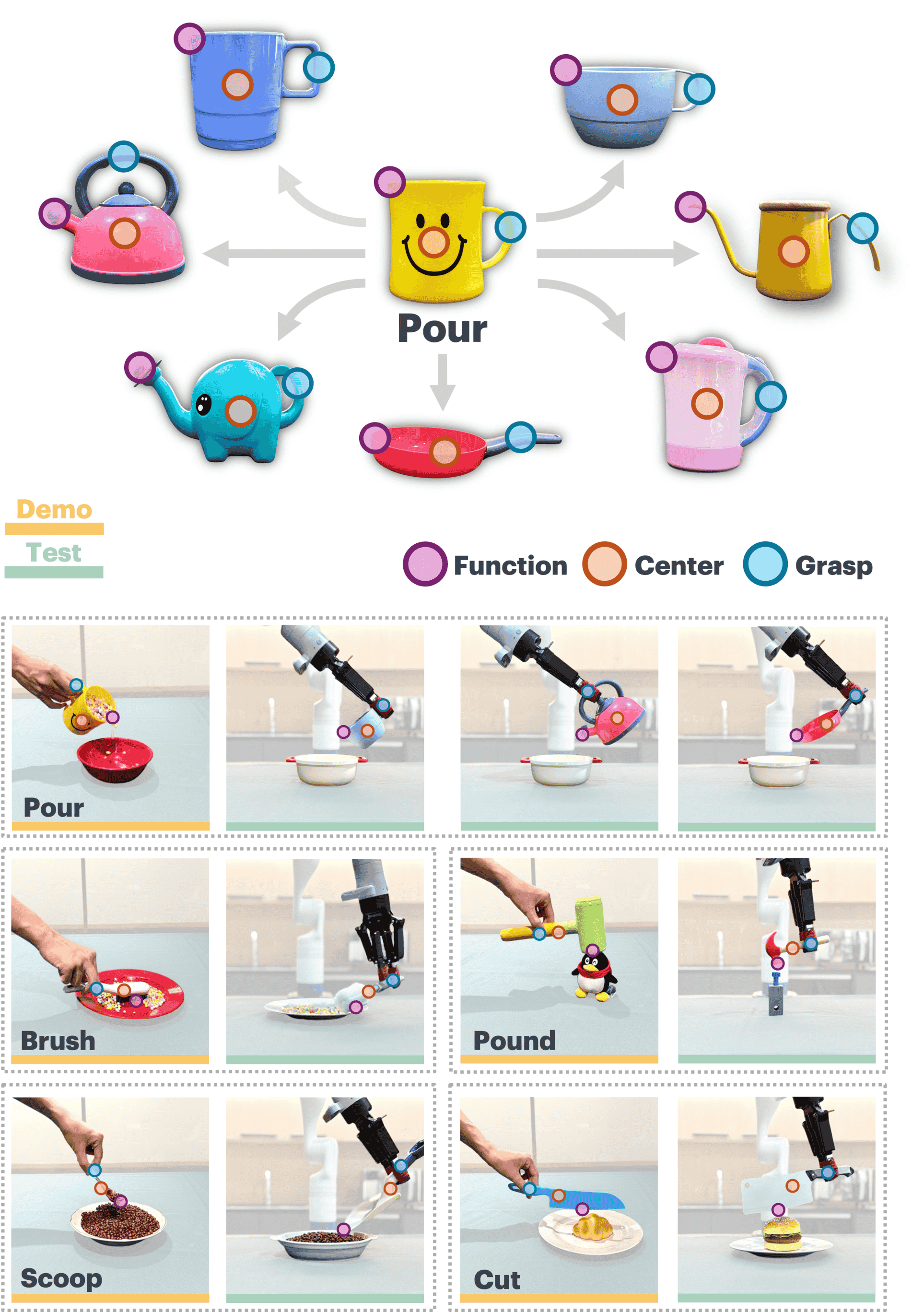}};
  \end{tikzpicture}
    \vspace*{-0.3in}
  \caption{FUNCTO establishes functional correspondences between demonstration and test tools using 3D functional keypoints. With a single human demonstration video, FUNCTO generalizes the demonstrated tool manipulation skill to novel tools, even with significant intra-function geometric variations.}
  \label{fig:concept}
  \vspace*{-0.8cm}
\end{figure}

While humans can effortlessly achieve the objective described above, it remains a non-trivial challenge for robots due to significant geometric variations (e.g., shape, size, topology) among tools supporting the same function (i.e., intra-function variations). 
As shown in Figure \ref{fig:concept}, although both the mug and the teapot support the same function of pouring,  their geometries differ significantly (e.g., the teapot features a long neck and a handle positioned on top of its body). 
Apparently, to successfully apply OSIL in this case, a key challenge lies in establishing functional correspondences between demonstration and test tools.
Previous OSIL methods \cite{vitiello2023one, heppert2024ditto, di2024dinobot, zhu2024vision, li2024okami, biza2023one, zhang2024one} assume that tools supporting the same function share highly similar shapes or appearances. 
Based on this assumption, they establish ``shallow" correspondences through techniques such as keypoint-based pose estimation \cite{vitiello2023one, heppert2024ditto, di2024dinobot}, global point set registration \cite{zhu2024vision, li2024okami}, shape warping \cite{biza2023one}, and invariant region matching \cite{zhang2024one} to align geometrically or visually similar tools. However, this assumption often fails in practice due to large intra-function variations.  
As a result, previous OSIL methods exhibit limited generalization to novel tools. This limitation motivates us to ask: What remains invariant among tools with the same function despite significant intra-function variations? Pioneering studies in cognitive anthropology \cite{washburn1960tools} reveal that humans exhibit highly consistent \textit{behavioral patterns} when using different tools serving the same purpose. For instance, the behavioral pattern of pouring involves approaching the tool (e.g., mug), grasping it, and directing its spout towards the target object (e.g., bowl). This spatiotemporal pattern remains invariant across tools (e.g., mug, teapot, saucepan) with the same function of pouring. 

Inspired by this observation, we propose FUNCTO (\underline{Func}tion-Centric OSIL for \underline{To}ol Manipulation), which emphasizes the functional aspects of tool correspondences over geometric or visual similarities as in previous works. FUNCTO achieves this by establishing function-centric correspondences between demonstration and test tools using a 3D functional keypoint representation. The 3D functional keypoint representation consists of a function point, where the tool interacts with the target (e.g., the spout of a mug); a grasp point, where the tool is held (e.g., the handle of a mug); and a center point, which is the tool's 3D center. By focusing on these three functional keypoints, FUNCTO captures the invariant spatiotemporal pattern among tools supporting the same function while ignoring function-irrelevant geometric details. Specifically, FUNCTO is factorized into three stages: (1) Functional keypoint extraction, which detects functional keypoints and tracks their motions from the human demonstration video; (2) Function-centric correspondence establishment, which transfers functional keypoints from the demonstration tool to the test tool and establishes function-centric correspondences using geometric constraints on the functional keypoints; (3) Functional keypoint-based action planning, which uses the demonstration and test functional keypoints to generate a robot end-effector trajectory for task execution.

We evaluate FUNCTO against existing OSIL methods and behavioral cloning (BC) methods through extensive real-robot experiments on diverse tool manipulation tasks. Leveraging the proposed function-centric approach with 3D functional keypoints, FUNCTO addresses the limitations of previous works that rely solely on geometric or visual similarities and achieves better generalization to novel tools with the same function despite significant intra-function variations. \\

\noindent \textbf{Contribution.} \ The main contribution of this work is a novel formulation of function-centric correspondence using a 3D functional keypoint representation for tool manipulation. This enables robots to generalize tool manipulation skills from a single human demonstration video to novel tools with the same function despite significant intra-function variations.

\section{Related Work}

\subsection{One-Shot Imitation Learning}
OSIL has been explored in various domains, such as image recognition \cite{vinyals2016matching, santoro2016meta}, generative models \cite{edwards2017towards}, and reinforcement learning \cite{duan2016rl}. The objective is to generalize the demonstrated behavior to novel instances or variations of the task with minimal prior knowledge or additional training.

In robotics, early OSIL works \cite{finn2017one, duan2017one, yu2018one} propose to leverage prior knowledge through meta-learning across a diverse set of robot tasks or skills. Following a ``pre-training and adapting" strategy, Wen et al. \cite{wen2022you, wen2022catgrasp} learn a category-level canonical representation during the pre-training stage and adapt it to new instances at inference. Similarly, Zhang et al. \cite{zhang2024one} conduct in-domain pre-training of a graph-based invariant region matching network and generalize to geometrically similar tools with a single demonstration. However, these methods face two major limitations: (1) they require in-domain pre-training, and (2) their generalization is restricted to geometrically or visually similar tools, struggling to handle out-of-domain tools with significant intra-function variations. Meanwhile, there has been a growing trend of using behavioral cloning (BC) models \cite{chi2023diffusion, zhao2023learning, ze20243d}, pre-trained on massive expert demonstrations, to generalize to unseen tools and configurations. We will later experimentally show that by clearly articulating the functional correspondences between tools, FUNCTO achieves better generalization performance compared to state-of-the-art BC methods, even with significantly less data.

Similar to our setup, \cite{heppert2024ditto} utilizes a transformer-based local feature matching model to compute the relative pose transformations between demonstration and test tools. Meanwhile, \cite{vitiello2023one} and \cite{di2024dinobot} leverage the off-the-shelf vision foundation model DINO \cite{oquabdinov2} to establish semantic correspondences between visually similar tools. In parallel, \cite{zhu2024vision} and \cite{li2024okami} employ global point set registration \cite{choi2015robust} to align demonstration and test tools. Similarly,  \cite{biza2023one} adopts shape warping \cite{rodriguez2018transferring} to correspond instances within the same category. Nevertheless, these methods assume that demonstration and test tools share highly similar shapes or appearances, limiting their generalization to novel tools with large geometric variations. In contrast, FUNCTO can handle significant intra-function variations to enable generalization to novel tools.

\subsection{Keypoint Representation for Tool Manipulation}
Keypoint representation has been extensively studied in tool manipulation \cite{manuelli2019kpam, manuelli2021keypoints, qin2020keto, xu2021affordance, liu2024moka, huangrekep, gao2023k, gao2024bi}, as it provides a compact and expressive way of encoding object information in terms of both semantics and actionability. For instance, KETO \cite{qin2020keto} introduces a task-specific keypoint generator trained with self-supervision for planar tool manipulation. KPAM \cite{manuelli2019kpam} uses 3D semantic keypoints as the tool representation to accomplish category-level manipulation tasks. Similarly, K-VIL \cite{gao2023k, gao2024bi} leverages a categorical correspondence model \cite{florence2018dense} to extract keypoint-based geometric constraints from one or few-shot human demonstrations.

More recent works leverage foundation models to predict semantic keypoints in zero-shot and generate corresponding tool manipulation motions. MOKA \cite{liu2024moka} generates planar manipulation motions via mark-based visual prompting \cite{nasirianypivot}. While FUNCTO also employs a similar technique for functional keypoint detection, it can predict tool trajectories in 3D, enabling it to handle more complex tool manipulation tasks. ReKep \cite{huangrekep} proposes to represent a manipulation task as a list of task-specific keypoint constraints and predict semantic keypoints in a zero-shot manner. However, these constraints require significant manual effort, and the zero-shot keypoint extraction strategy is highly error-prone. In contrast, FUNCTO effortlessly extracts tool manipulation constraints and functional keypoints from human demonstration videos and does not require any object/task-specific knowledge to define the constraints. Furthermore, compared to the zero-shot keypoint proposal as in \cite{liu2024moka, huangrekep}, human demonstrations provide valuable cues for keypoint localization. Leveraging these cues enhances task performance, as demonstrated in the experiment section.

\subsection{Visual Correspondence in Robotics}
The ability to establish correspondences \cite{zhang2024tale} between seen and unseen scenarios is essential for robots to generalize. Techniques from the computer vision community, such as pose estimation \cite{wen2024foundationpose}, optical flow estimation \cite{bailer2017cnn}, and point tracking \cite{karaev2025cotracker}, have been widely adopted in robotic tasks and applications. In robotic manipulation, DON \cite{florence2018dense} learns dense visual correspondences with self-supervision for transferring grasps across visually similar object instances. Building on this concept, TransGrasp \cite{wen2022transgrasp} adopts Deformed Implicit Field \cite{deng2021deformed} to build shape correspondences within the same object category for grasp transfer. More recently, NDF \cite{simeonov2022neural} proposes a neural implicit representation to learn categorical descriptors from few-shot demonstrations. While these approaches focus on building visual correspondences within the same category, FUNCTO extends this capability to establishing functional correspondences, even in the presence of significant intra-function variations. 

FUNCTO is also closely related to affordance theory \cite{gibson1977theory} and the functional correspondence problem \cite{lai2021functional}. A common principle shared by FUNCTO and these works is that correspondences should extend beyond geometric or visual similarity to incorporate functional relevance.

\begin{figure*}[t]
  \centering
  \vspace*{-0.2in}
  \begin{tikzpicture}[inner sep = 0pt, outer sep = 0pt]
    \node[anchor=south west] (fnC) at (0in,0in)
      {\includegraphics[height=5.5in,clip=true,trim=0in 0in 0in 0in]{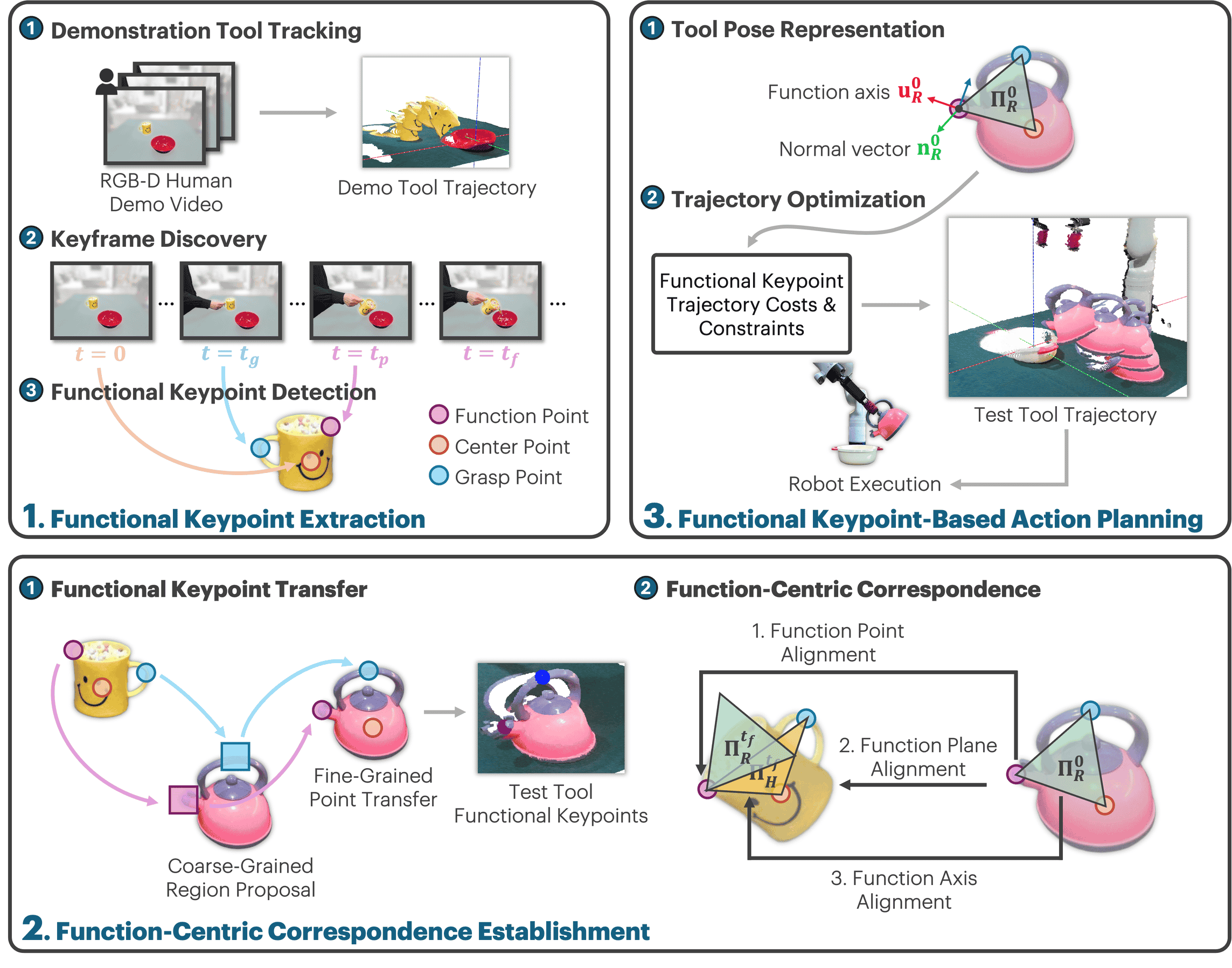}};
  \end{tikzpicture}
    \vspace*{-0.1in}
  \caption{An overview of the FUNCTO framework. The pipeline consists of three stages: (1) Functional keypoint extraction, where functional keypoints and their trajectories are extracted from the human demonstration video; (2) Function-centric correspondence establishment, where function-centric correspondences between demonstration and test tools are established using geometric constraints on the functional keypoints; and (3) Functional keypoint-based action planning, where the test tool trajectory is synthesized and executed to accomplish a functionally equivalent task.}
  \label{fig:pipeline}
  \vspace*{-0.2in}
\end{figure*}

\section{Problem Formulation}
We consider the problem of enabling the robot to imitate the tool manipulation behavior demonstrated in a single human video to accomplish functionally equivalent tasks using novel tools with the same function. Specifically, each task involves a robot grasping a tool (object) with a parallel-jaw gripper to interact with a target (object) in a tabletop environment. The task is defined by a list of spatiotemporal constraints between the tool and the target.



During the demonstration phase, a human performs a tool manipulation task, recording a sequence of RGB-D images, $\mathcal{V}_H = \{I_t\}_{t=0}^{N-1}$, with a stationary camera, where $N$ denotes a finite task horizon. The sequence $\mathcal{V}_H$ is paired with a natural language task description $l_H$ (e.g., \textit{``use the \underline{mug} to \underline{pour} contents into the \underline{bowl}"}) that specifies three elements: a tool (e.g., \textit{mug}), a target (e.g., \textit{bowl}), and a function (e.g., \textit{pour}). During inference, given a robot observation $o_R$ and a corresponding task description $l_R$, the objective is to develop an OSIL policy $\pi$, mapping $o_R$ and $l_R$ to a robot end-effector trajectory $\tau_R = \{a_t\}_{t=0}^{N-1}$ that maximizes the likelihood of task success. Here, $a_t = (R_t, T_t) \in \text{SE(3)}$ represents the 6-DoF end-effector pose at timestep $t$, where $R_t \in \text{SO(3)}$ and $T_t \in \mathbb{R}^3$ denote 3D orientation and translation, respectively. 

\noindent \textbf{Assumptions.} During the implementation, we have made the following assumptions: (1) Visual observations are single-view and do not contain any action annotations. (2) The robot has no object/task-specific prior knowledge, such as 3D object models or manual task constraints, but has access to commonsense knowledge embedded in foundation models. (3) No in-domain pre-training is conducted. (4) Tools are modeled as rigid objects and can be manipulated by the designated gripper.

\section{FUNCTO}

\subsection{Overview}

In this section, we describe FUNCTO, a function-centric OSIL framework for tool manipulation. FUNCTO consists of three stages: (1) functional keypoint extraction, (2) function-centric correspondence establishment, and (3) functional keypoint-based action planning. An overview of the proposed framework is presented in Figure \ref{fig:pipeline}. Specifically, in the first stage (Section \ref{detection}), FUNCTO detects 3D functional keypoints $K_H$ and track their motions $\{K_H^t\}_{t=0}^{N-1}$ from $\mathcal{V}_H$. In the second stage (Section \ref{correspondence}), $K_H$ are transferred from the demonstration tool to the test tool to obtain test functional keypoints $K_R$. Subsequently, FUNCTO establishes function-centric correspondences with $K_H$ and $K_R$. In the final stage (Section \ref{planning}), FUNCTO generates a robot end-effector trajectory $\tau_R$, using the reference $\{K_H^t\}_{t=0}^{N-1}$ and the established function-centric correspondences, for execution.

\subsection{Functional Keypoint Extraction}\label{detection}

\noindent \textbf{Demonstration Tool Tracking.} FUNCTO starts by localizing and segmenting the tool and the target in the first frame of $\mathcal{V}_H$ (i.e., $I_0$) using Grounding-SAM \cite{ren2024grounded}. Subsequently, $N_k$ keypoints are uniformly sampled within the tool mask. We use Cotracker \cite{karaev2025cotracker} to capture their 3D motions (using the depth information) for the rest of $\mathcal{V}_H$, yielding their 3D trajectories in the camera frame. To ensure the extracted motion is independent of the absolute locations of the tool and target, FUNCTO transforms the 3D keypoint trajectories from the camera frame to the target (object) frame by estimating the target object pose in the camera frame. The origin of the target frame, denoted as the target point $p_{\text{target}}$, is defined as the 3D center of the target. Finally, FUNCTO utilizes rigid body transformation \cite{bottema1990theoretical} to calculate the relative transformations of the tool between consecutive timesteps based on the 3D keypoint trajectories. Throughout this section, all 3D elements are represented in the target frame unless otherwise specified.  \\


\noindent \textbf{Keyframe Discovery.} Due to the partial observability of functional keypoints in the human demonstration video, FUNCTO discovers keyframes where functional keypoints on the demonstration tool can be effectively identified. Specifically, FUNCTO discovers four keyframes: (1) the initial keyframe $I_0$ ($t=0$), where the tool is in its initial state; (2) the grasping keyframe $I_g$ ($t=t_g$), where the hand grasps the tool; (3) the function keyframe $I_f$ ($t=t_f$), where the interaction between the tool and target starts; (4) the pre-function keyframe $I_{p}$ ($t=t_{p}$), where the interaction is about to start while both the tool and target remain clearly visible. These keyframes satisfy $0 < t_g < t_p < t_f < N-1$. Note that $I_{p}$ is essential for function point detection before heavy occlusion associated with the interaction occurs. For detecting $I_g$, we use an off-the-shelf hand-object detector from \cite{shan2020understanding}. Following \cite{zhu2024vision}, an unsupervised change point detection method \cite{killick2012optimal} is employed to identify $I_f$ based on velocity statistics derived from the 3D keypoint trajectories. Lastly, we backtrack through $\mathcal{V}_H$ to locate a preceding frame (i.e., $I_{p}$) where the occlusion between the tool and the target, measured by IoU, is below a pre-defined threshold. Examples of discovered keyframes are presented in Figure \ref{fig:pipeline} (stage 1).\\

\noindent \textbf{Functional Keypoint Detection.} Once the keyframes are identified, FUNCTO proceeds to detect 3D functional keypoints and track their motions, denoted as $\{K_H^t\}_{t=0}^{N-1} = \{[p_{\text{func}}^t, p_{\text{grasp}}^t, p_{\text{center}}^t]\}_{t=0}^{N-1}$. $p_{\text{func}}^t$, $p_{\text{grasp}}^t$, and $p_{\text{center}}^t$ represent the 3D locations of the function point, grasp point, and center point at timestep $t$, respectively, with $p \in \mathbb{R}^3$.

Three functional keypoints are detected in the respective keyframes. The grasp point is determined by computing the intersection between the hand mask and the tool mask in $I_g$, while the center point is computed as the 3D center of the tool in $I_0$. Detecting the function point is non-trivial, as there may be no physical contact between the tool and the target (e.g., pouring), and it also requires commonsense knowledge about tool usage. Therefore, FUNCTO leverages the multi-modal reasoning capabilities of VLMs and employs mark-based visual prompting \cite{nasirianypivot}. The visual prompting process involves two steps: (1) task-agnostic meta-prompt definition and (2) task-specific prompting. In the first step, we provide a meta-prompt with task-agnostic context information to the VLM, including the definition of the function point, the expected behavior of the VLM, and the desired response format. In the second step, we sample and annotate $N_c$ candidate points on the tool's boundary in $I_{p}$ with Farthest Point Sampling \cite{qi2017pointnet}, assigning each point a unique index. Guided by visual cues from the human demonstration indicating where interaction occurs, $I_{p}$ and $l_H$ are then used to prompt the VLM with a multi-choice problem among $N_c$ candidates to determine the function point.

\begin{figure}[t]
  \centering
  \vspace*{-0.2in}
  \begin{tikzpicture}[inner sep = 0pt, outer sep = 0pt]
    \node[anchor=south west] (fnC) at (0in,0in)
      {\includegraphics[height=1.5in,clip=true,trim=0.2in 0in 0in 0in]{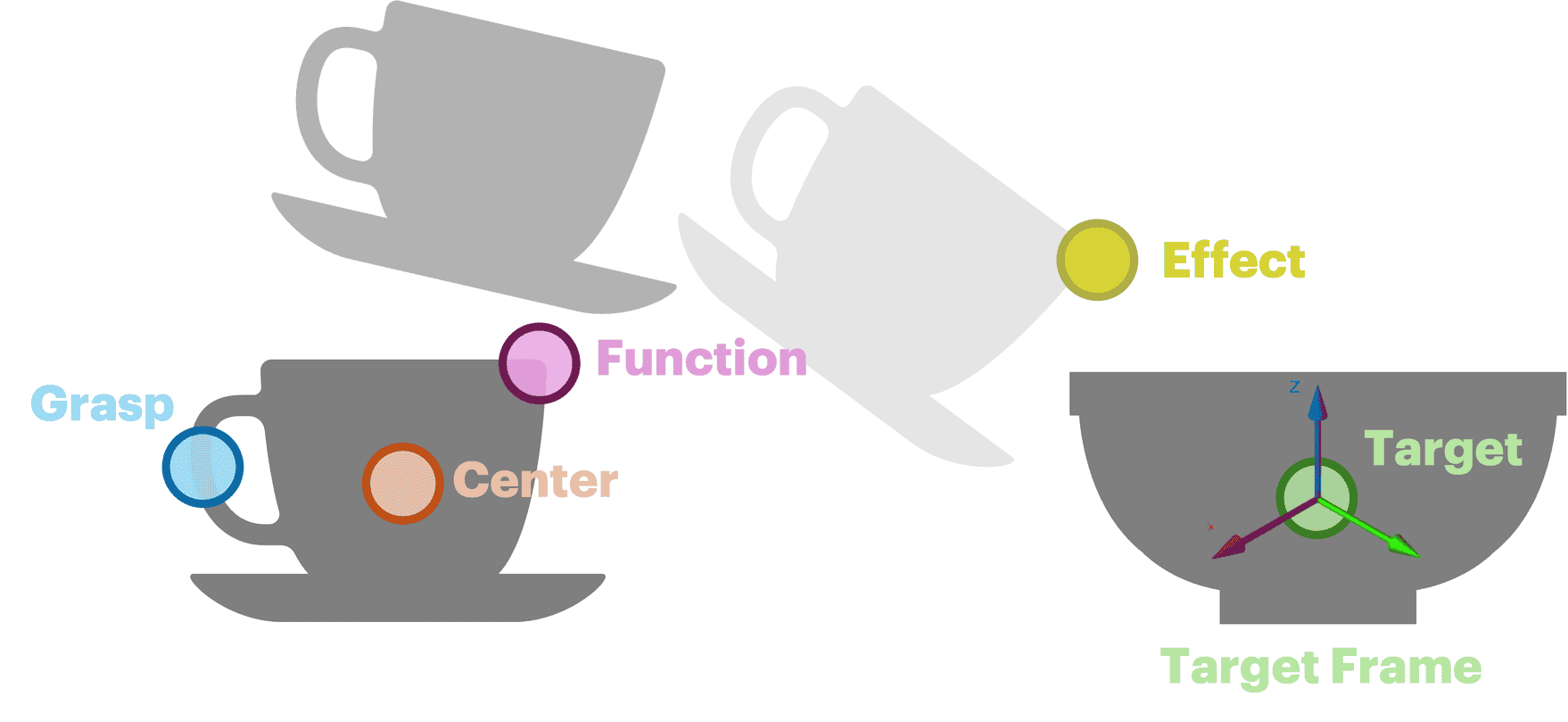}};
  \end{tikzpicture}
    \vspace*{-0.2in}
  \caption{A graphical illustration of the function point, grasp point, center point, effect point, target point, and target frame.}
  \label{fig:point_frame_def}
  \vspace*{-0.2in}
\end{figure}

After identifying three functional keypoints in their respective keyframes, we track their motions using the previously computed sequence of relative transformations, resulting in $\{K_H^t\}_{t=0}^{N-1}$. Additionally, $p_{\text{func}}^{t_f}$ is attached to the target, referred to as the effect point $p_{\text{eff}}$, to represent the 3D location where the interaction occurs on the target object. Figure \ref{fig:point_frame_def} provides a graphical illustration of the function point, grasp point, center point, effect point, target point, and target frame.

\subsection{Function-Centric Correspondence Establishment}\label{correspondence}
\noindent \textbf{Functional Keypoint Transfer.} At the core of FUNCTO lies the function-centric correspondence establishment using functional keypoints. FUNCTO achieves this by first transferring the functional keypoints from the demonstration tool to the test tool, obtaining $K_R^0 = [q_{\text{func}}^0, q_{\text{grasp}}^0, q_{\text{center}}^0]$, where $q \in \mathbb{R}^3$. 

\begin{figure}[t]
  \centering
  \hspace*{-0.1in}
  \begin{tikzpicture}[inner sep = 0pt, outer sep = 0pt]
    \node[anchor=south west] (fnC) at (0in,0in)
      {\includegraphics[height=1.0in,clip=true,trim=0in 0in 0in 0in]{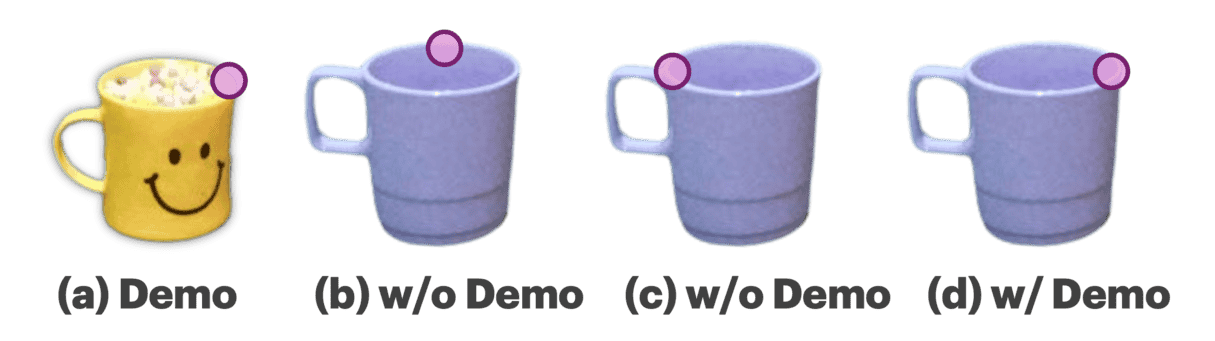}};
  \end{tikzpicture}
    \vspace*{-0.3in}
  \caption{Qualitative results of function point transfer. (a) shows the function point extracted from the human demonstration. Function points in (b) and (c) are proposed by the VLM in a zero-shot manner. (d) shows the transferred function point using (a) as a reference.}
  \label{fig:keypoint_proposal}
  \vspace*{-0.2in}
\end{figure}

The functional keypoint transfer process, illustrated in Figure \ref{fig:pipeline} (stage 2),  consists of a coarse-grained region proposal and a fine-grained point transfer. In the first step, we begin by projecting $p_{\text{func}}^0$ and $p_{\text{grasp}}^0$ from the 3D space onto the image plane $I_0$. The marked $I_0$ serves as a reference for identifying test tool functional keypoints. Providing such a reference is essential, as the functional keypoint location is coupled with the demonstrated human action. When the test tool has multiple possible functional keypoints, selecting a functional keypoint not matching the intended action can lead to failure. Figure \ref{fig:keypoint_proposal} provides a qualitative comparison of function point transfer with and without using the reference.

Similar to Section \ref{detection}, FUNCTO first utilizes mark-based visual prompting to propose coarse candidate regions on the test tool for 2D function and grasp points. Compared to functional keypoint detection in the previous stage, two key differences are: (1) the marked  $I_0$  is additionally provided to the VLM as a reference, and (2) the selected candidate point is expanded into a candidate region, with its size adaptively adjusted based on the 2D dimensions of the test tool. In the second step, we employ a pre-trained dense semantic correspondence model \cite{zhang2024tale} to precisely transfer 2D function and grasp points from $I_0$ to candidate regions in $o_R$, resulting in  $q_{\text{func}}^0$  and $q_{\text{grasp}}^0$. The dense semantic correspondence model provides finer point-level correspondences compared to using the VLM alone. $q_{\text{center}}^0$ is computed as the 3D center of the test tool. For the target object, we scale $p_{\text{eff}}$  based on the 3D dimension ratio between the demonstration and test targets to obtain  $q_{\text{eff}}$. $K_R$ and  $q_{\text{eff}}$  are expressed in the test target frame. \\



\noindent \textbf{Function-Centric Correspondence.} FUNCTO formulates the function-centric correspondence as geometric constraints on 3D functional keypoints, inspired by \cite{manuelli2019kpam}. Specifically, this step specifies the function keyframe constraint (i.e., the desired test tool state at $t_f$) for trajectory generation.

\begin{figure}[t]
  \centering
  \hspace*{-0.3in}
  \begin{tikzpicture}[inner sep = 0pt, outer sep = 0pt]
    \node[anchor=south west] (fnC) at (0in,0in)
      {\includegraphics[height=1.2in,clip=true,trim=0in 0in 0in 0in]{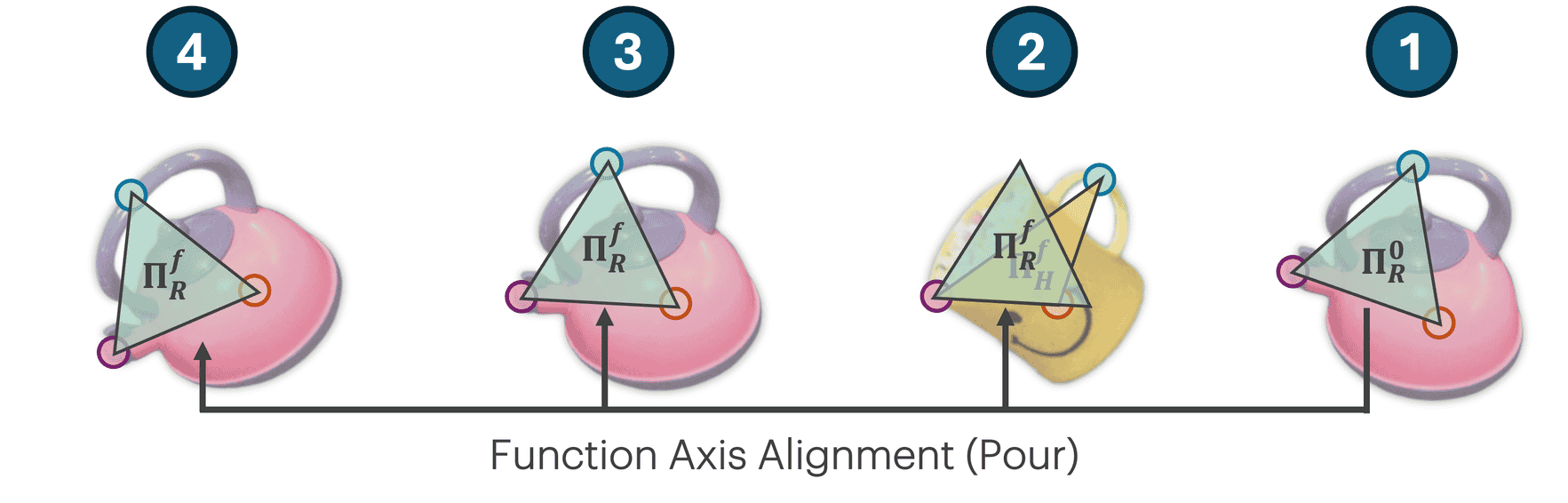}};
  \end{tikzpicture}
    \vspace*{-0.3in}
  \caption{An illustration of function axis alignment process: (1) test function plane \( \Pi_R^0 \), (2) demonstration function plane  \( \Pi_H^{t_f} \), (3) initially aligned test function plane  \( \Pi_R^{t_f} \), and (4) VLM refined test function plane  \( \Pi_R^{t_f} \).}
  \label{fig:func_axis_align}
  \vspace*{-0.25in}
\end{figure}

Formally, the function keyframe constraint can be represented by a rigid transformation $\mathbf{T}_{\text{func}} \in \text{SE(3)}$ that aligns $K_H^{t_f}$ and $K_R^0$. This process, illustrated in Figure \ref{fig:pipeline} (stage 2), can be divided into the following steps:
\begin{enumerate}[label=\arabic*., start=0]
    \item \textbf{Function Plane Construction}: Given $K_H^{t_f}$ and $K_R^0$, function planes \( \Pi_H^{t_f} \) and \( \Pi_R^0 \) are constructed as follows:
\begin{itemize}
    \item \(\mathbf{u}_H^{t_f}\) (function axis): A normalized vector pointing from the center to the function point at $t_f$.
    \item \(\mathbf{v}_H^{t_f}\): A normalized vector pointing from the function to the grasp point at $t_f$.
    \item \(\mathbf{n}_H^{t_f}\): The unit normal vector at $t_f$.
\end{itemize}
Similarly, \(\mathbf{u}_R^0\), \(\mathbf{v}_R^0\), and \(\mathbf{n}_R^0\) are defined for \( \Pi_R^0 \). \\

    \item \textbf{Function Point Alignment}: The function points should be aligned to ensure that the interaction occurs at the desired location of the test tool. The function point alignment is defined by the following constraint:
    \begin{align*}
        \| \mathbf{T}_{\text{point}} \begin{bmatrix} q_{\text{func}}^0 \\ 1 \end{bmatrix} - \begin{bmatrix} p_{\text{func}}^{t_f} \\ 1 \end{bmatrix} \| = 0
    \end{align*}  

    \item \textbf{Function Plane Alignment}: The normal vectors should be aligned to ensure that function planes have the same orientation. The function plane alignment is defined by the following constraint:
    \begin{align*}                
    \mathbf{n}_H^{t_f} \cdot (\text{rot}(\mathbf{T}_{\text{plane}})\mathbf{n}_R^0) = 1
    \end{align*}
    where $\text{rot}(\mathbf{T})$ denotes the rotation component of a rigid transformation $\mathbf{T}$. \\
    
    \item \textbf{Function Axis Alignment}: The function axes, which are function-relevant operational vectors, should be aligned to ensure that the test tool is properly tilted relative to the target (e.g., pitch angle for pouring). The function axis alignment is defined by the following constraint:
    \begin{align*}    
    \mathbf{u}_H^{t_f} \cdot (\text{rot}(\mathbf{T}_{\text{axis}}) \mathbf{u}_R^0) = 1
    \end{align*}
    However, due to structural differences between the demonstration and test tools  (i.e., differences in relative locations of functional keypoints), directly applying $\mathbf{T}_{\text{axis}}$ to the test tool may not result in successful task execution. For instance, a mug and a teapot may require different pouring angles. To address this issue, $\mathbf{T}_{\text{axis}}$ is further refined using the VLM. Specifically, a pre-defined set of angle offsets, ranging from $[-45^{\circ}, -45^{\circ}]$, is applied to $\mathbf{T}_{\text{axis}}$. For each offset, the combined point cloud of the test tool and target is back-projected onto the camera plane. The VLM is then prompted to identify the rendered image that represents the optimal state conducive to the task success, given the demonstration function keyframe $I_f$ as a reference. The transformation corresponding to the optimal state is recorded as  $\mathbf{T}^{*}_{\text{axis}}$. Qualitative results of function axis alignment are illustrated in Figure \ref{fig:func_axis_align}.
\end{enumerate}
Finally, the geometric constraints from each step are combined to compute  $\mathbf{T}_{\text{func}}$ and $K_R^{t_f}$ :
\begin{align*}
\mathbf{T}_{\text{func}} = \mathbf{T}_{\text{point}} \cdot \mathbf{T}_{\text{plane}} \cdot \mathbf{T}^{*}_{\text{axis}}, \
K_R^{t_f} = \mathbf{T}_{\text{func}} K_R^0
\end{align*}
To ensure that the function point interacts with the effect point at $t_f$, $q_{\text{func}}^{t_f}$ is further adjusted to align with  $q_{\text{eff}}$. Meanwhile, the same adjustment is applied to  $q_{\text{grasp}}^{t_f}$ and  $q_{\text{center}}^{t_f}$. The resulting $K_R^{t_f}$ represents the predicted test tool state at $t_f$.


\subsection{Functional Keypoint-Based Action Planning}\label{planning}
\noindent \textbf{Tool Pose Representation.} In the final stage, FUNCTO computes the test tool pose at each timestep and generates a robot end-effector trajectory $\tau_R$ for task execution. Specifically, the demonstration tool pose at timestep $t$ can be represented, using $K_H^t$, as:
\begin{align*}
     \mathbf{T}_H^t =
\begin{bmatrix}
\mathbf{R}_H^t & p_{\text{func}}^t \\
\mathbf{0} & 1
\end{bmatrix}\end{align*}
where $\mathbf{R}_H^t$ is the rotation matrix derived from the function axis $\mathbf{u}_H^t$ and the normal vector $ \mathbf{n}_H^t$. With such a pose representation, demonstration functional keypoint trajectory $\{K_H^t\}_{t=0}^{N-1}$ can be transformed to a sequence of $\text{SE(3)}$ poses $\{\mathbf{T}_H^t\}_{t=0}^{N-1}$. Similarly, the test tool pose at $t$ is represented as:
\begin{align*}
\mathbf{T}_R^t =
\begin{bmatrix}
\mathbf{R}_R^t & q_{\text{func}}^t \\
\mathbf{0} & 1
\end{bmatrix}
\end{align*}
where $\mathbf{R}_R^t$ is similarly defined. An example of the tool pose representation is illustrated in Figure \ref{fig:pipeline} (stage 3). The function keyframe state $K_R^{t_f}$ and the initial keyframe state $K_R^{0}$ are transformed to the pose representations $\mathbf{T}_{\text{func}}$ and $\mathbf{T}_{\text{init}}$, respectively. \\







\noindent \textbf{Tool Trajectory Optimization.} Given the function keyframe pose $\mathbf{T}_{\text{func}}$, the initial keyframe pose $\mathbf{T}_{\text{init}}$, and the reference pose trajectory $\{\mathbf{T}_H^t\}_{t=0}^{N-1}$, the optimization problem for solving the test tool trajectory $\{\mathbf{T}_R^t\}_{t=0}^{N-1}$ can be formulated as:
\begin{align*}
\min_{\{\mathbf{T}_R^t \}_{t=0}^{N-1}} & \sum_{t=0}^{N-1} \left(
\| q_{\text{func}}^t - p_{\text{func}}^t \|_2^2  + \| \text{Log}(\mathbf{R}_R^t (\mathbf{R}_H^t)^\top)\|^2 \right) \\
\text{s.t.}  \quad & \mathbf{T}_R^0  = \mathbf{T}_{\text{init}} \\
 \quad & \mathbf{T}_R^{t_f}  = \mathbf{T}_{\text{func}}
\end{align*}
where $\text{Log}:\text{SO(3)} \rightarrow \mathbb{R}^3$\cite{sola2018micro}. This formulation can flexibly incorporate additional costs and constraints, such as smoothness costs and collision avoidance constraints. Implementation details can be found in Appendix E. \\

\noindent \textbf{Tool Trajectory Execution.} 
The test tool trajectory in test target frame $\{\mathbf{T}_R^t\}_{t=0}^{N-1}$ is first transformed into the robot base frame $\{\mathbf{T}_{R_\text{base}}^t\}_{t=0}^{N-1}$. Then, we use GraspGPT \cite{tang2023graspgpt} to sample a 6-DoF task-oriented grasp pose around $q_{\text{grasp}}^0$ on the test tool. Assuming the gripper is rigidly attached to the test tool after grasping, the robot end-effector trajectory $\tau_R$ can be computed with the sampled grasp pose and $\{ \mathbf{T}_{R_\text{base}}^t\}_{t=0}^{N-1}$.
This trajectory is tracked and executed using operational space control.






\begin{figure*}[t]
  \vspace*{-0.2in}
  \hspace*{-0.3in} 
  \begin{tikzpicture}[inner sep = 0pt, outer sep = 0pt]
    \node[anchor=south west] (fnC) at (-0in, 0in)
      {\includegraphics[height=2.3in,clip=true,trim=0in 0in 0in 0in]{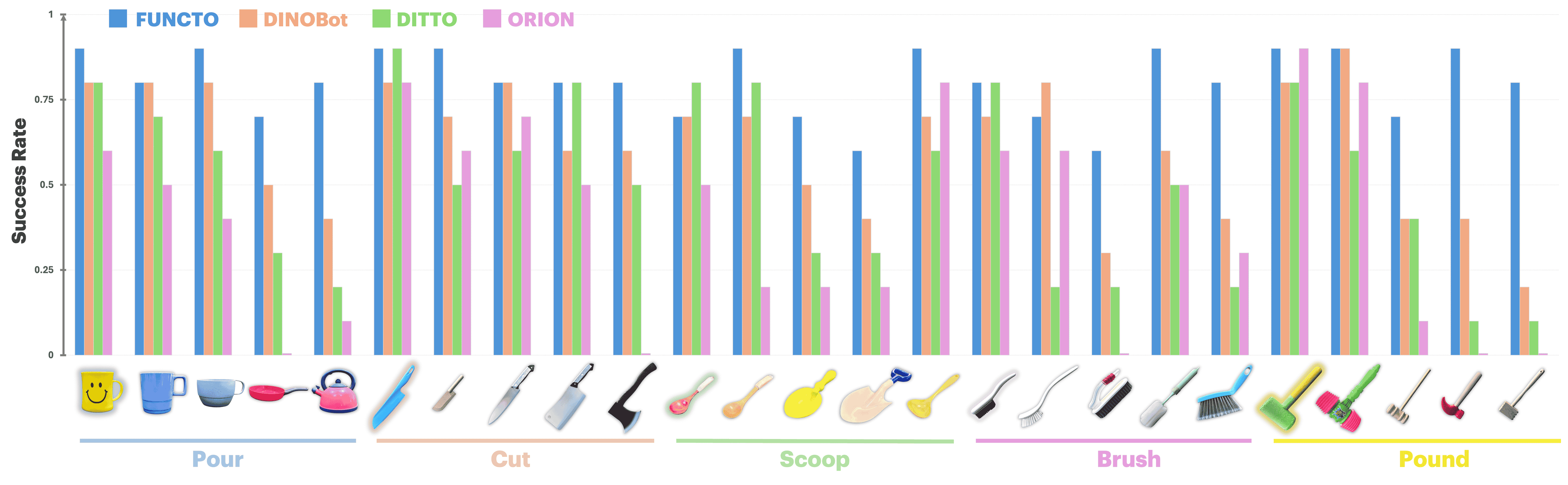}};
  \end{tikzpicture}
    \vspace*{-0.3in}
  \caption{Quantitative comparison to one-shot imitation learning baselines. The first tool of each function (highlighted) is used for demonstration.}
  \label{fig:osil_exp}
\end{figure*}

\begin{table*}[th] \small
\centering
\caption{Quantitative comparison to Behavioral Cloning baselines}
  \vspace*{-0.1in}
\renewcommand\arraystretch{1.8}
\setlength\tabcolsep{3pt}
\begin{tabular}{ccclcclcclcclcclcc}
\toprule
\multirow{2}{*}{\textbf{Method}} & \multicolumn{2}{c}{\textbf{Pour}} &  & \multicolumn{2}{c}{\textbf{Cut}} &  & \multicolumn{2}{c}{\textbf{Scoop}} &  & \multicolumn{2}{c}{\textbf{Brush}} &  & \multicolumn{2}{c}{\textbf{Pound}} &  & \multicolumn{2}{c}{Overall}                    \\ \cline{2-3} \cline{5-6} \cline{8-9} \cline{11-12} \cline{14-15} \cline{17-18} 
                                 & Seen           & Unseen           &  & Seen           & Unseen          &  & Seen            & Unseen           &  & Seen            & Unseen           &  & Seen            & Unseen           &  & \multicolumn{1}{l}{Seen} & \multicolumn{1}{l}{Unseen} \\ \midrule
ACT (50 Demos)                   & 60.0\%             & 17.5\%                 &  & 70.0\%             & 40.0\%                &  & \textbf{70.0\%}                & 47.5\%                 &  & 50.0\%                & 32.5\%                 &  & 50.0\%                & 25.0\%                &  & 60.0\%                        & 32.5\%                           \\
DP (50 Demos)                    & 50.0\%              & 20.0\%                &  & 50.0\%             & 22.5\%                &  & 40.0\%                & 27.5\%                &  & 50.0\%               & 35.0\%                 &  & 40.0\%               & 27.5\%                &  & 57.50\%                        & 26.50\%                            \\
DP3 (50 Demos)                   & 40.0\%              & 20.0\%                 &  & 40.0\%               & 25.0\%               &  & 20.0\%                & 15.0\%                 &  & 20.0\%               & 12.5\%                &  & 40.0\%              & 15.0\%                 &  & 32.00\%                          & 17.50\%                            \\ 
FUNCTO (1 Demo)                  & \textbf{90.0\%}               & \textbf{80.0\%}                 &  & \textbf{90.0\%}               & \textbf{82.5\%}                &  & \textbf{70.0\%}                & \textbf{77.5\%}                 &  & \textbf{80.0\%}                & \textbf{75.0\%}                 &  & \textbf{90.0\%}                & \textbf{82.5\%}                 &  & \textbf{84.00\%}                          & \textbf{79.50\%}                            \\ \bottomrule
\end{tabular}
  \vspace*{-0.2in}
\label{tab:bc_exp}
\end{table*}

\begin{figure*}[t]
  \centering
  \vspace*{-0.2in}
  \hspace*{-0.1in} 
  \begin{tikzpicture}[inner sep = 0pt, outer sep = 0pt]
    \node[anchor=south west] (fnC) at (-0in, 0in)
      {\includegraphics[height=5.6in,clip=true,trim=0in 0in 0in 0in]{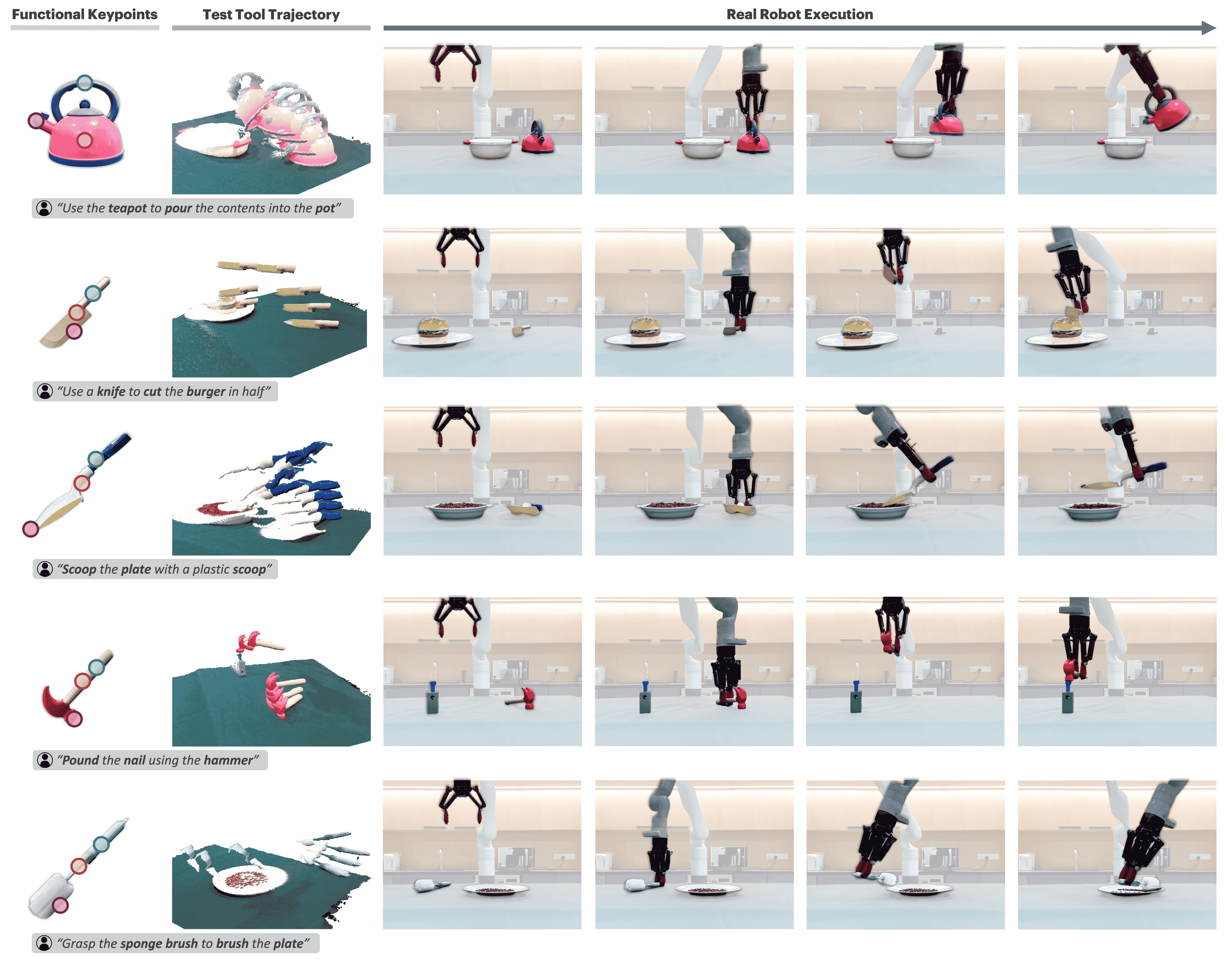}};
  \end{tikzpicture}
    \vspace*{-0.2in}
  \caption{Qualitative results of predicted functional keypoints, test tool trajectories, and real-robot executions across five functions.}
  \label{fig:qualitative}
  \vspace*{-0.2in}
\end{figure*}

\section{Experiments}
In this section, we conduct real-robot experiments to validate FUNCTO's effectiveness and analyze key design choices. Specifically, we answer the following questions: (1) How well does FUNCTO generalize from a single human demonstration to novel tools? (2) How does FUNCTO perform compared to existing OSIL methods under the same setup? (3) How does FUNCTO compare to state-of-the-art BC methods? (4) How does each design choice in function-centric correspondence establishment (Section \ref{correspondence}) affect the overall performance?

\subsection{Experimental Setup}

\noindent \textbf{Baselines.} We compare FUNCTO against the following OSIL baselines: (1) \textbf{\textsc{DinoBot}} \cite{vitiello2023one, di2024dinobot}, which leverages the visual correspondence capability of the vision foundation model DINO to perform semantic feature extraction and correspondence. (2) \textbf{\textsc{Ditto}} \cite{heppert2024ditto}, which employs a pre-trained visual correspondence model LOFTR to estimate the relative pose transformations between demonstration and test tools. (3) \textbf{\textsc{Orion}} \cite{zhu2024vision, li2024okami}, which extracts geometric features with Fast-Point Feature Histograms and performs a global-local registration. We adopt the original correspondence implementations of these baselines. The low-level execution components remain consistent with FUNCTO. OSIL methods with different setups, such as those requiring in-domain pre-training or object/task-specific prior knowledge, are excluded for a fair comparison.

In addition to OSIL baselines, we also compare FUNCTO with BC baselines, which represent more typical imitation learning approaches in recent works. Specifically, the following BC baselines are compared: (1) \textbf{\textsc{Action Chunking Transformer}} (ACT) \cite{zhao2023learning}, a transformer-based BC method with action chunking and temporal ensemble. (2) \textbf{\textsc{Diffusion Policy}} (DP) \cite{chi2023diffusion}, a diffusion-based BC method that models a visuomotor policy as a conditional denoising diffusion process. For both ACT and DP, we use the pre-trained DINO-ViT as the backbone for visual feature extraction. (3) \textbf{\textsc{3D Diffusion Policy}} (DP3) \cite{ze20243d}, a more recent diffusion-based BC method that operates on 3D visual representations extracted from sparse point clouds with a lightweight point encoder. \\

\noindent \textbf{Task Description.} We evaluate FUNCTO and baselines on five tool manipulation functions: \texttt{pour}, \texttt{cut}, \texttt{scoop}, \texttt{brush}, and \texttt{pound}. A tool manipulation task is defined by pairing a function with a tool and a target  (e.g., \texttt{mug-pour-bowl}). In this work, we primarily focus on addressing intra-function variations between tools, placing less emphasis on target variations. For each function, we design five tasks using different tools, divided into three levels of generalization: (1) spatial generalization (seen), where the demonstration tool is randomly positioned in the workspace; (2) instance generalization (unseen), where the demonstration and test tools are different instances from the same category; (3) category generalization (unseen), where demonstration and test tools are from different categories with the same function. In the context of this paper, ``same function" refers to imitating the tool manipulation behavior demonstrated by the human to accomplish a functionally equivalent task. 

\noindent \textbf{Experimental Protocol.} During the demonstration phase, a single-view, actionless video of a human performing a tool manipulation task is recorded with a stationary RGB-D camera, accompanied by a task description, for each function. During the testing phase, an RGB-D image of the workspace is captured using a similar camera setup and sent to the robot for action planning and execution. For training the BC baselines, we collect 50 human demonstrations with teleportation for each function. For testing the BC baselines, we use the same targets as those in the demonstrations to emphasize the impact of variations in the test tools. In terms of performance evaluation, each method is tested with 10 trials per task, resulting in a total of 50 trials per function and 250 trials across all five functions. The detailed task success conditions are described in Appendix B. The average success rate is used as the evaluation metric.

\subsection{Experimental Results}
\noindent \textbf{Quantitative Comparison to OSIL Baselines.} The detailed quantitative evaluation results are reported in Figure \ref{fig:osil_exp}. Each function is evaluated with five tasks. The first task tests spatial generalization, the next two tasks evaluate instance generalization, and the final two tasks assess category generalization. 

All methods perform well in spatial generalization with the seen demonstration tool, achieving success rates above 70\%. However, all baselines exhibit significant performance drops (from 20\% to 40\%) when generalizing to novel tool instances and categories, especially for those with substantial differences in shape, size, or topology. For instance, in \texttt{teapot-pour-pot}, the teapot and the demonstrated mug differ in both shape and part topology. The teapot has a conical body, a long neck, and a handle positioned on top, whereas the mug features a cylindrical body with a side-mounted handle. We also observe that variations in size and scale negatively affect the performance of the baselines. In \texttt{hammer-pound-nail}, the red hammer is much smaller than the demonstrated mallet, causing inaccurate pounding point alignment in 3D. This results in infeasible contact between the tool and the target, highlighting the significance of the proposed function-centric correspondence. 

Among all baselines, ORION relies solely on geometric features, rendering it ineffective at handling large geometric variations. DINOBot outperforms both DITTO and ORION, achieving an average success rate of 57.5\% when generalizing to novel tools. This performance can be attributed to DINO's strong visual correspondence capability. However, DINOBot still struggles to establish correspondences between visually distinct tools due to intra-function variations. In contrast, FUNCTO significantly outperforms the OSIL baselines, achieving a high success rate of 79.5\% across five functions for novel tool generalization. Figure \ref{fig:qualitative} visualizes the qualitative results of real-robot executions across five functions. \\


\noindent \textbf{Quantitative Comparison to BC Baselines.} The quantitative evaluation results are summarized in Table \ref{tab:bc_exp}. We divide the results into Seen and Unseen categories to emphasize the performance gap between demonstration and test tools. For seen tools, BC baselines trained on 50 demonstrations exhibit some level of generalization to different spatial layouts, with the two leading baselines, ACT and DP, achieving success rates ranging from 50\% to 60\%.  By modeling the relative spatial relationship between the tool and the target, FUNCTO inherently supports spatial generalization.  

However, all BC baselines struggle with unseen tool generalization, primarily due to intra-function variations, with success rates approximately half those of the Seen category. In contrast, FUNCTO achieves a significantly higher success rate of 79.5\%, attributed to the proposed function-centric correspondence. We also evaluate BC baselines trained with 10 and 1 demonstration(s). However, they fail to produce meaningful results and are therefore excluded from the report. \\

\begin{figure}[t]
  \centering
  \begin{tikzpicture}[inner sep = 0pt, outer sep = 0pt]
    \node[anchor=south west] (fnC) at (0in,0in)
      {\includegraphics[height=1.5in,clip=true,trim=0in 0in 1in 0in]{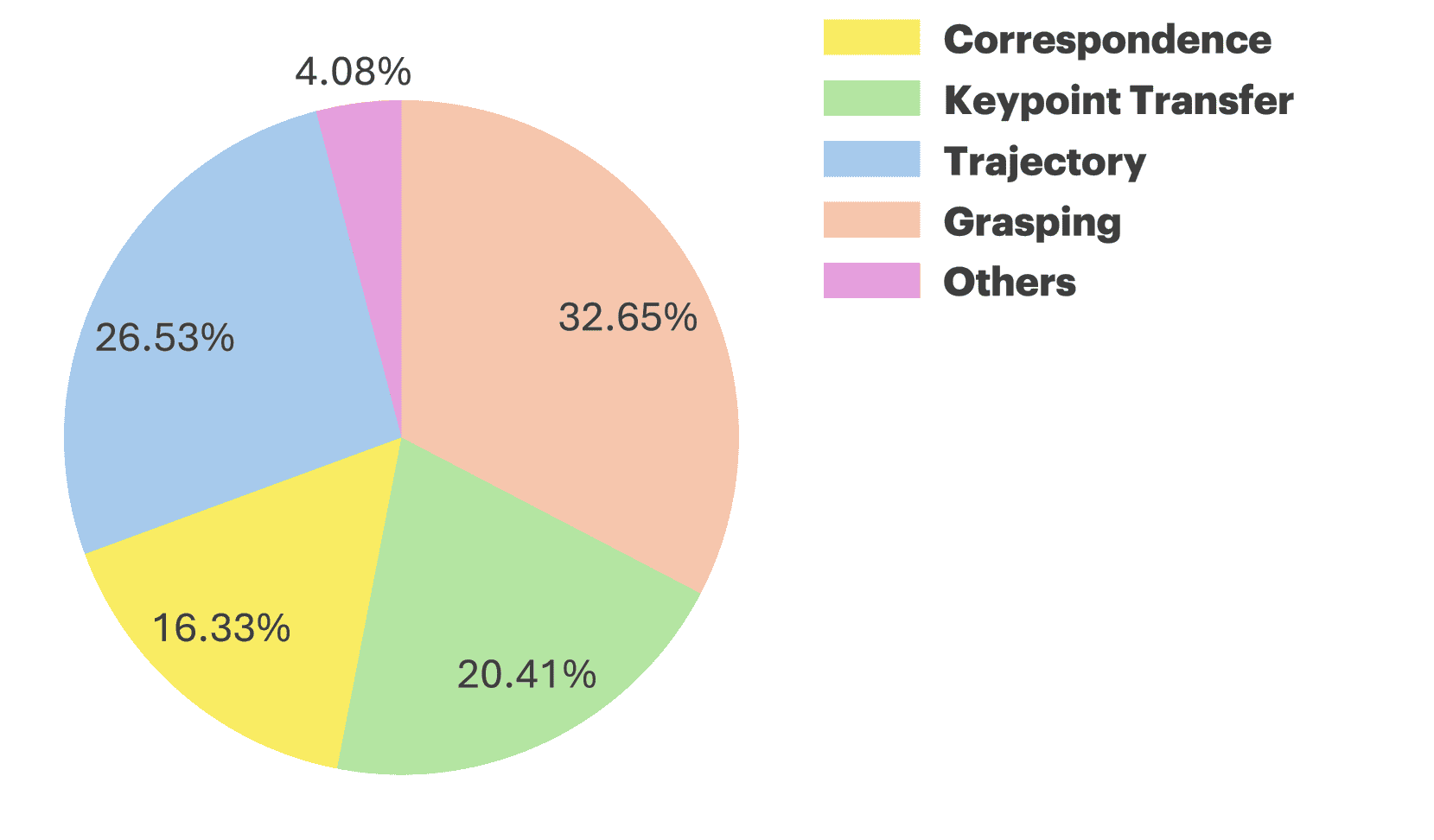}};
  \end{tikzpicture}
    \vspace*{-0.1in}
  \caption{Failure analysis of system components}
  \label{fig:error}
  \vspace*{-0.3in}
\end{figure}

\noindent \textbf{Failure Analysis.} The modular design of FUNCTO facilitates the interpretation and in-depth analysis of failure cases. The result of the failure analysis is reported in Figure \ref{fig:error}. The identified failure sources are categorized into: (1) function-centric correspondence, (2) functional keypoint transfer, (3) trajectory planning, (4) grasping, and (5) others (e.g., segmentation, detection). 

The primary failures arise from (4) and (3). Specifically, failures in grasping often occur when the tool flips or slips due to unstable contact between the tool and the gripper, preventing the robot from completing the task. Incorporating an online state tracking module (probably using multiple calibrated cameras) and closed-loop execution could potentially mitigate this issue. For trajectory planning, failures primarily result from unexpected collisions between the tool and the target, particularly in contact-rich tasks (e.g., \texttt{scrubber-brush-plate}). Providing visual-tactile feedback is essential for successfully accomplishing such tasks. Functional keypoint transfer errors are mainly caused by incorrect candidate region proposals for function points but contribute less significantly to overall failures. These errors may be mitigated as VLMs continue to improve. Correspondence errors are mainly attributed to inaccurate depth information of the functional keypoints. Empirically, the function-centric correspondence works well with accurate 3D functional keypoint locations.

\subsection{Ablation study}
To gain further insights into the design choices behind the core component of FUNCTO, function-centric correspondence establishment, we conduct two sets of ablation studies: one on functional keypoint transfer and another on function-centric correspondence. Performance is evaluated on five tool manipulation tasks: \texttt{teapot}-\texttt{pour}-\texttt{pot}, \texttt{knife}-\texttt{cut}-\texttt{burger}, \texttt{hammer}-\texttt{pound}-\texttt{nail}, \texttt{scoop}-\texttt{scoop}-\texttt{bowl}, \texttt{scrubber}-\texttt{brush}-\texttt{plate}. \\

\begin{figure*}[t]
  \hspace*{0in}
  \vspace*{-0.1in}
  \begin{tikzpicture}[inner sep = 0pt, outer sep = 0pt]
    \node[anchor=south west] (fnC) at (0in,0in)
      {\includegraphics[height=2.3in,clip=true,trim=0in 0in 0in 0in]{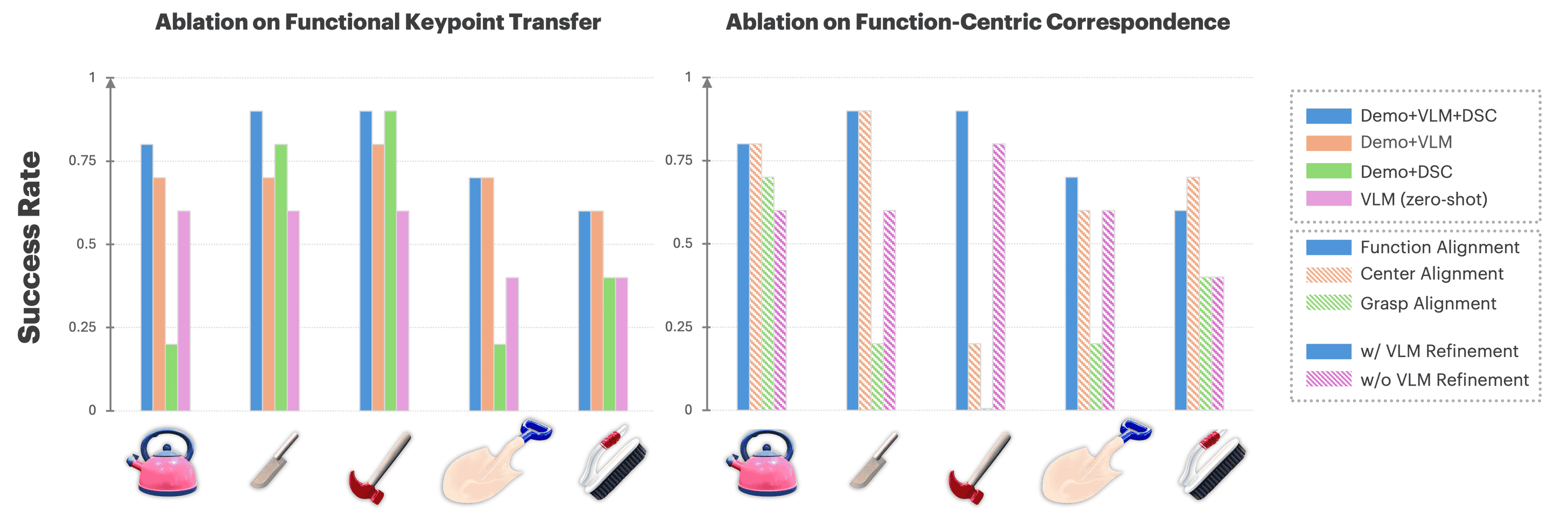}};
  \end{tikzpicture}
  \caption{Ablation studies on functional keypoint transfer (left) and function-centric correspondence (right).}
  \label{fig:ablation}
  \vspace*{-0.2in}
\end{figure*}

\noindent \textbf{Ablation on Functional Keypoint Transfer.} We evaluate four functional keypoint transfer strategies: (1) Demo+VLM+DSC (proposed), which utilizes demonstration functional keypoints as references to prompt the VLM for region proposal, followed by point transfer through a dense semantic correspondence model; (2) Demo+VLM, which removes the dense semantic correspondence model from the proposed implementation; (3) Demo+DSC, which relies solely on a dense semantic correspondence model for functional keypoint transfer; (4) VLM (zero-shot), which directly prompts the VLM to propose functional keypoints in a zero-shot manner, as in MOKA \cite{liu2024moka} and ReKep \cite{huangrekep}. The quantitative results are reported in Figure \ref{fig:ablation} (left). The proposed Demo+VLM+DSC consistently outperforms ablated versions. Demo+VLM performs reasonably well, benefiting from the rich commonsense knowledge embedded in VLMs. However, as indicated in \cite{rahmanzadehgervi2024vision}, VLMs struggle to provide precise point-level correspondences, particularly for tasks requiring high precision (e.g., \texttt{hammer}-\texttt{pound}-\texttt{nail}). On the other hand, solely relying on the dense semantic correspondence model often fails when dealing with large intra-function variations. The performance gap between Demo+VLM and VLM (zero-shot) justifies the point that demonstration functional keypoints provide valuable references for test functional keypoint proposal. Additional quantitative and qualitative evaluations are provided in Appendix C.\\

\noindent \textbf{Ablation on Function-Centric Correspondence.} In this ablation study, we investigate two questions: (1) Is aligning the function point more effective than aligning other functional keypoints? (2) Is VLM refinement necessary for function axis alignment? As shown in Figure \ref{fig:ablation} (right), function point alignment achieves the optimal performance in most cases by ensuring interactions occur at the desired location of the tool, regardless of variations in shape, topology, or size. When comparing strategies with and without VLM refinement, the latter rigidly aligns the function axes, ignoring changes in the relative locations of the three functional keypoints. This strategy may produce infeasible function keyframe poses. Incorporating commonsense knowledge from VLMs for function axis refinement statistically improves the performance. 




\section{Conclusion and Limitations}
\noindent \textbf{Conclusion.}  In this work, we present FUNCTO, a function-centric one-shot imitation learning framework for tool manipulation. At the core of FUNCTO is the idea of functional correspondence using a 3D functional keypoint representation. With such a formulation, FUNCTO generalizes the tool manipulation skill from a single human demonstration video to novel tools with the same function despite significant intra-function variations. Extensive real-robot experiments validate the effectiveness of FUNCTO, outperforming both modular one-shot imitation learning methods and end-to-end behavioral cloning methods. \\

\noindent \textbf{Limitations.}  Despite the promising results, several limitations remain: (1) Functional keypoint visibility and collinearity. Our current implementation assumes that the function point is clearly visible from the camera view. However, this assumption may not always hold, especially when learning from egocentric videos. Additionally, FUNCTO fails when the three functional keypoints are collinear in 3D. That said, such cases are uncommon for everyday tools. (2) State tracking and closed-loop execution. The current pipeline operates in an open-loop manner, which is sensitive to unexpected state changes or external disturbances. Integrating a state tracking module (probably using multiple calibrated cameras) and enabling closed-loop execution would further improve the robustness. (3) Multi-modal function modeling. Functions inherently exhibit multi-modality. For example, the function point on the mug shown in Figure \ref{fig:concept} is used for forward pouring, while points positioned on the left or right sides of the rim can facilitate side pouring. Although FUNCTO is currently limited to imitating a single usage of the function based on a single demonstration, future work will focus on capturing the multi-modality of a function from few-shot human demonstrations with a human-robot interaction system \cite{xiao2024robi}. Further discussions can be found in Appendix H.

\bibliographystyle{unsrt}
\bibliography{references}

\clearpage
\onecolumn

\section{Appendix}
\noindent \textbf{A. FUNCTO} \\
In this section, we provide additional qualitative results of the function-centric correspondences established by FUNCTO across five functions.

\begin{figure}[h]
  \centering
  \begin{tikzpicture}[inner sep = 0pt, outer sep = 0pt]
    \node[anchor=south west] (fnC) at (0in,0in)
      {\includegraphics[height=7in,clip=true,trim=0in 0in 0in 0in]{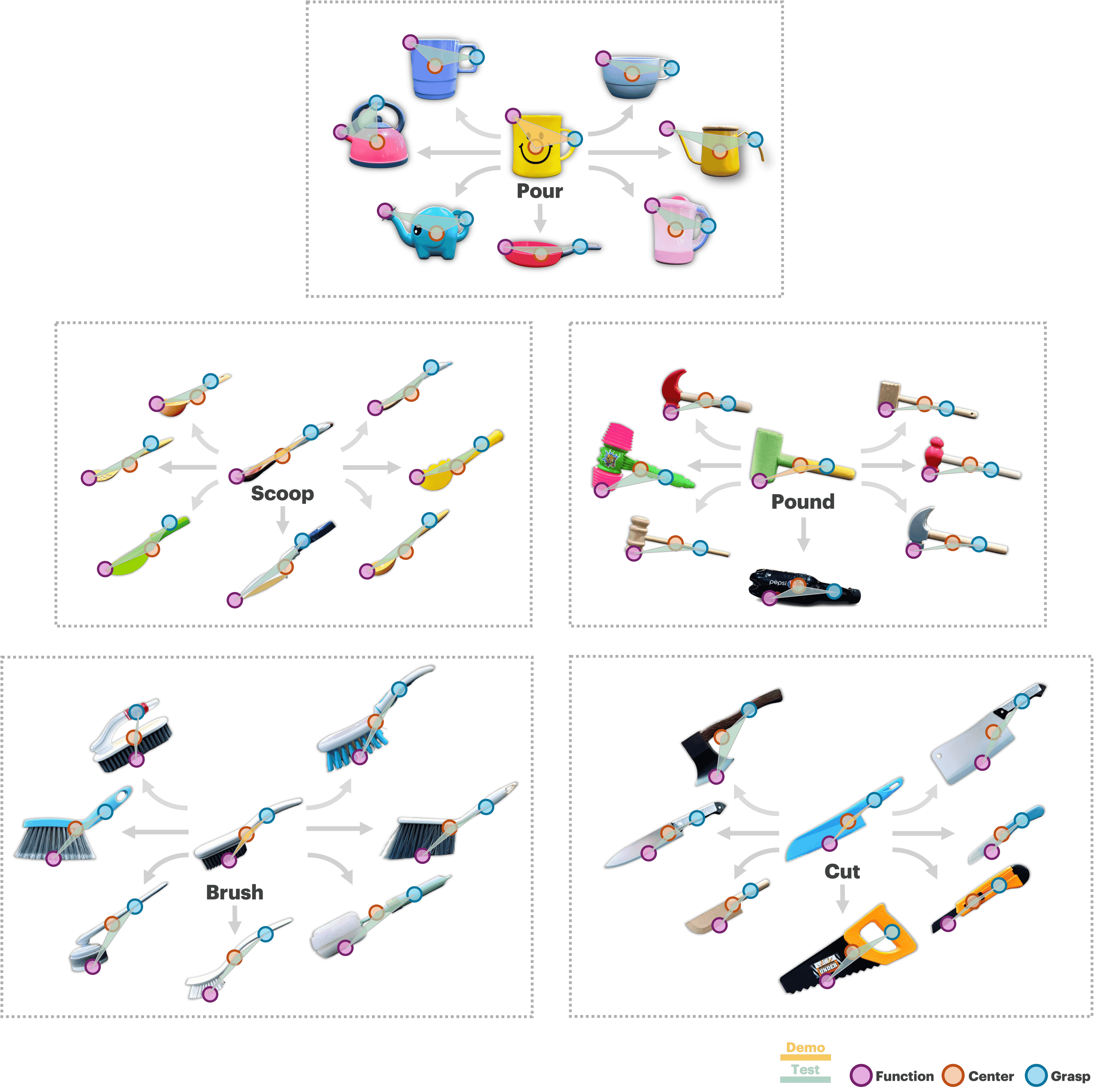}};
  \end{tikzpicture}
  \caption{Function-centric correspondences established by FUNCTO across five functions.}
  \label{fig:functos}
\end{figure}


\newpage

\noindent \textbf{B. Real-Robot Experiment}

\begin{figure}[h]
  \centering
  \vspace*{-0.1in}
  \begin{tikzpicture}[inner sep = 0pt, outer sep = 0pt]
    \node[anchor=south west] (fnC) at (0in,0in)
      {\includegraphics[height=2.3in,clip=true,trim=0in 0in 0in 0in]{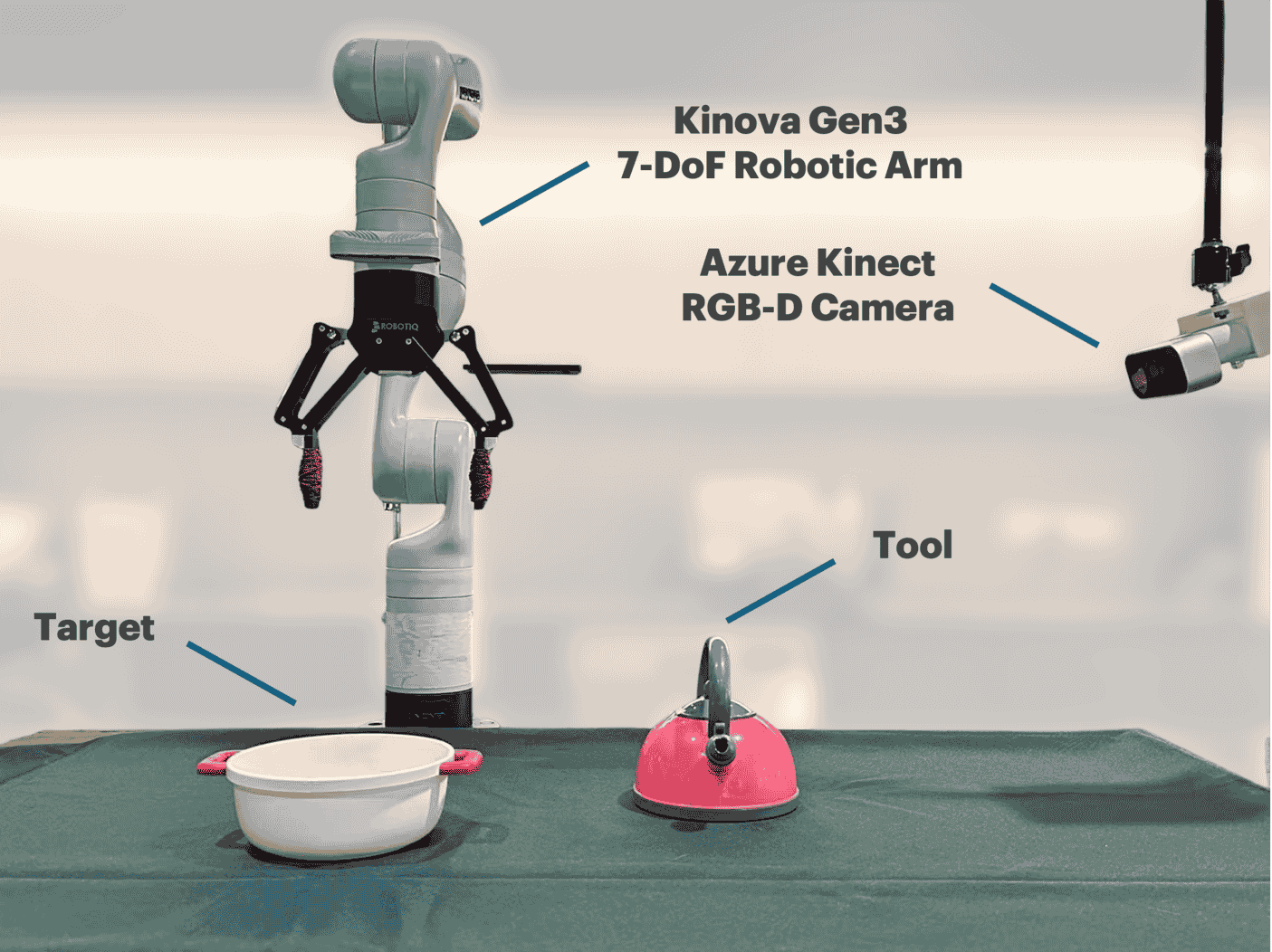}};
  \end{tikzpicture}
    \vspace*{-0.1in}
  \caption{Experimental platform}
  \label{fig:exp_setup}
\end{figure}

\noindent \textbf{Experimental Setup.} All experiments are conducted on the platform depicted in Figure \ref{fig:exp_setup}. The platform consists of a Kinova Gen3 7-DoF robotic arm and an Azure Kinect RGB-D camera. During each trial, a tool object and a target object are placed within the robot’s workspace, which is a 50cm × 30cm region.


\begin{figure}[h]
  \centering
  \begin{tikzpicture}[inner sep = 0pt, outer sep = 0pt]
    \node[anchor=south west] (fnC) at (0in,0in)
      {\includegraphics[height=3in,clip=true,trim=0in 0in 0in 0in]{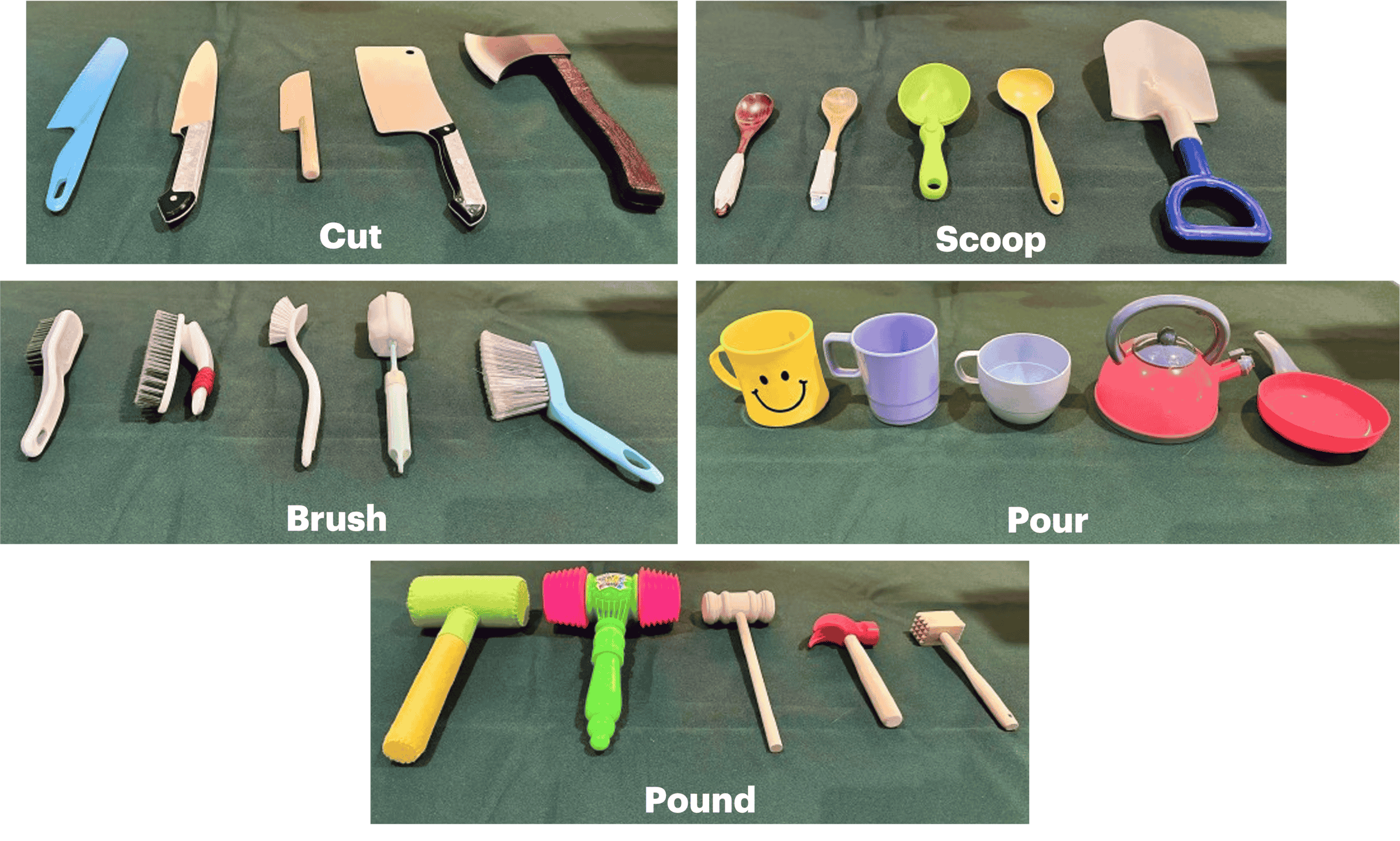}};
  \end{tikzpicture}
    \vspace*{-0.1in}
  \caption{Tool objects used in the real-robot experiments.}
  \label{fig:exp_tool}
\end{figure}

\begin{figure}[h]
  \centering
  \begin{tikzpicture}[inner sep = 0pt, outer sep = 0pt]
    \node[anchor=south west] (fnC) at (0in,0in)
      {\includegraphics[height=1.2in,clip=true,trim=0in 0in 0in 0in]{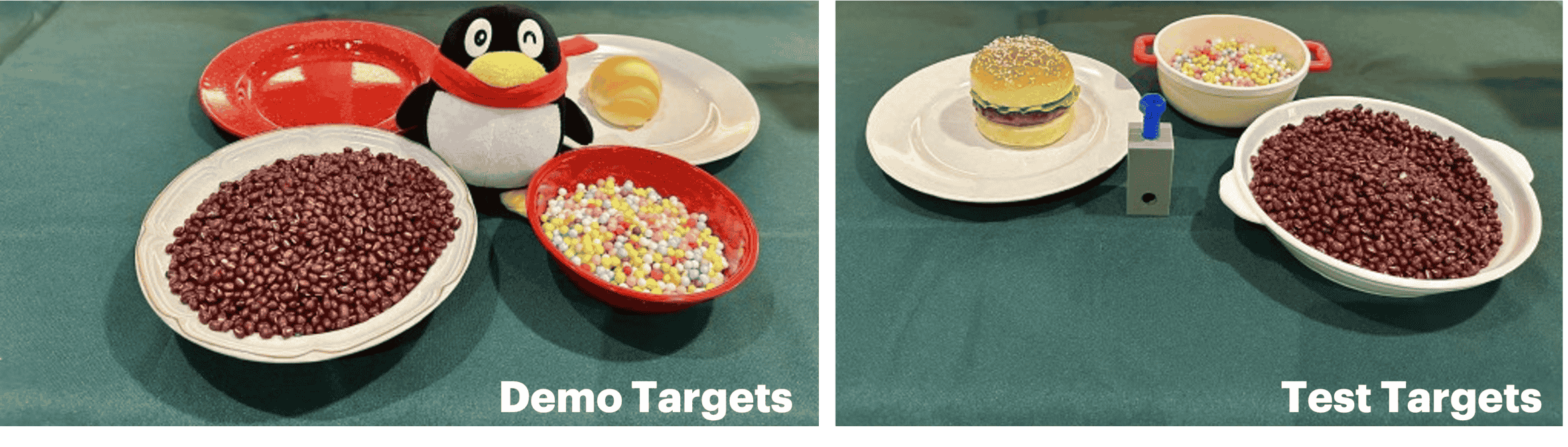}};
  \end{tikzpicture}
    \vspace*{-0.1in}
  \caption{Target objects used in the real-robot experiments.}
  \label{fig:exp_target}
\end{figure}

\noindent The tool and target objects used in the experiments are shown in Figure \ref{fig:exp_tool} and Figure \ref{fig:exp_target}, respectively. The leftmost tool of each function is used for human demonstration.

\newpage

\noindent \textbf{Task Success Conditions.} 

\begin{itemize}
    \item \textbf{Pour}: The particles within the tool are transferred into the target container.
    \item \textbf{Cut}: The blade of the tool makes contact with the target from above.
    \item \textbf{Scoop}: The tool collects and securely holds particles from the target container.
    \item \textbf{Pound}: The bottom of the tool head strikes the nail head.
    \item \textbf{Brush}: The tool moves across the target’s surface, displacing particles with its bristle. \\
\end{itemize}

\noindent \textbf{Qualitative Results.} 

\begin{figure}[h]
  \centering
  \begin{tikzpicture}[inner sep = 0pt, outer sep = 0pt]
    \node[anchor=south west] (fnC) at (0in,0in)
      {\includegraphics[height=6.8in,clip=true,trim=0in 0in 0in 0in]{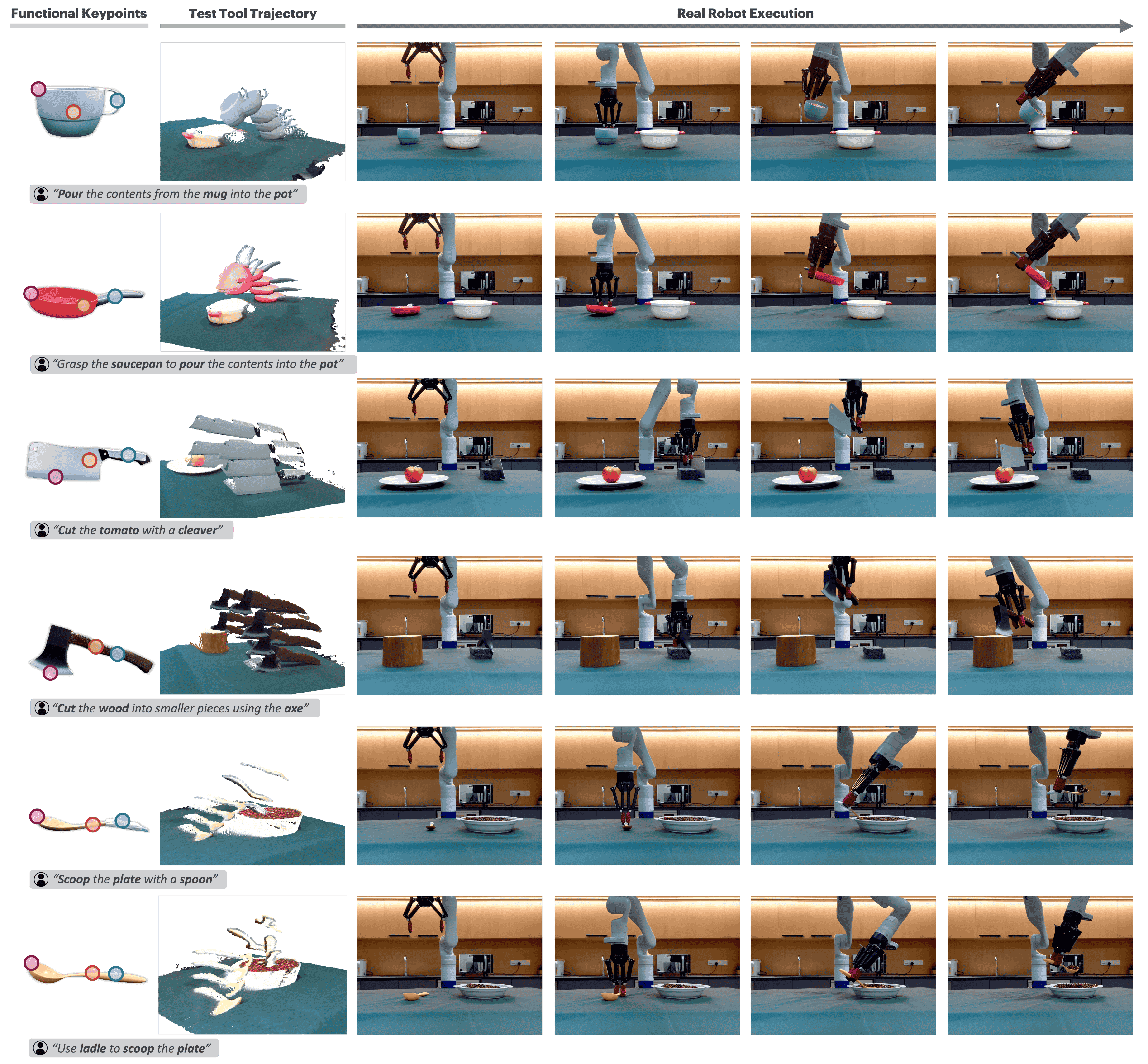}};
  \end{tikzpicture}
    \vspace*{-0.2in}
  \caption{Qualitative results of predicted functional keypoints, trajectories, and real-robot executions (pour, cut, scoop).}
  \label{fig:qualitative_1}
\end{figure}

\newpage

\begin{figure}[h]
  \centering
  \begin{tikzpicture}[inner sep = 0pt, outer sep = 0pt]
    \node[anchor=south west] (fnC) at (0in,0in)
      {\includegraphics[height=4.6in,clip=true,trim=0in 0in 0in 0in]{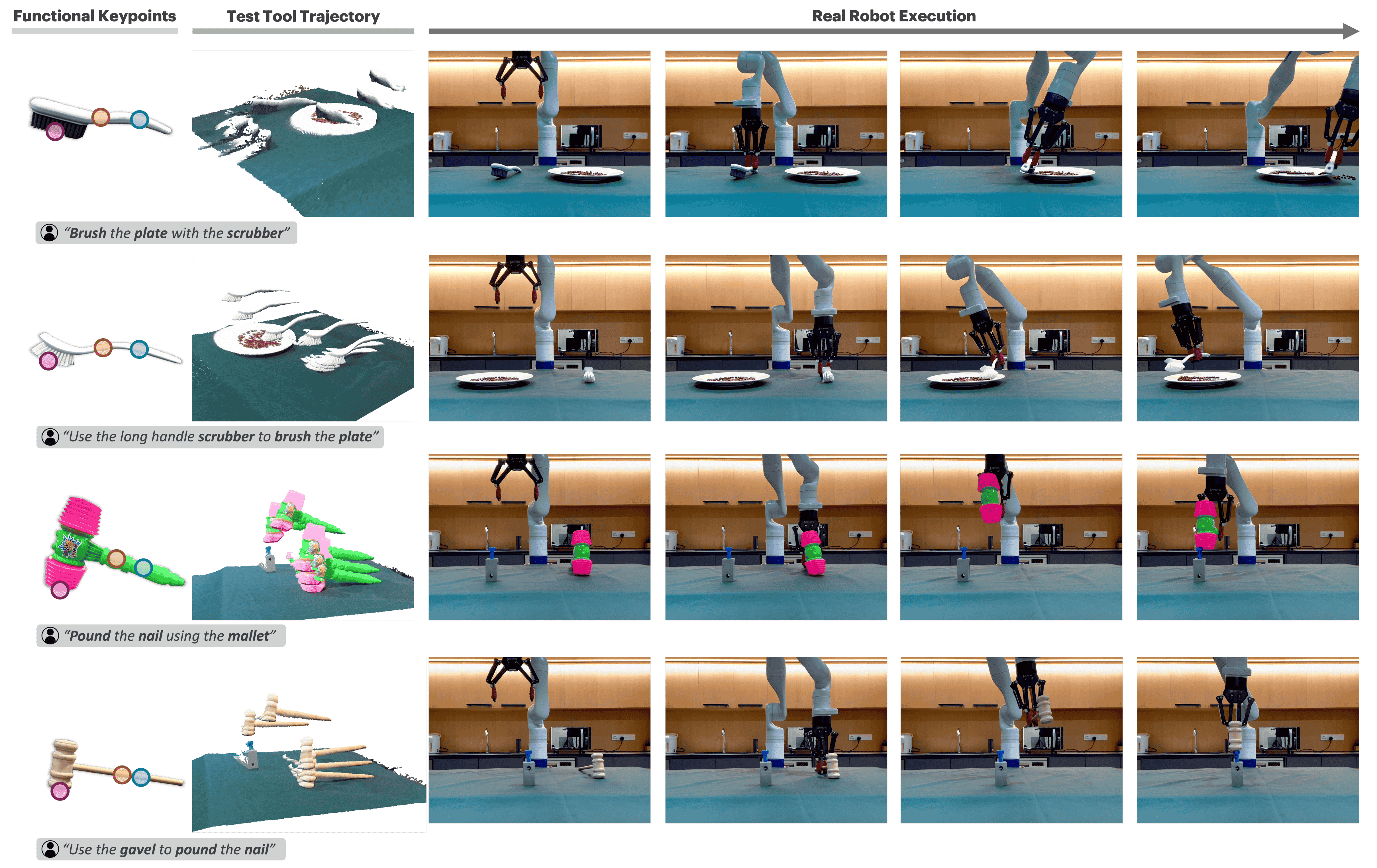}};
  \end{tikzpicture}
    \vspace*{-0.2in}
  \caption{Qualitative results of predicted functional keypoints, trajectories, and real-robot executions (brush, pound).}
  \label{fig:qualitative_2}
\end{figure}



\newpage

\noindent \textbf{C. Functional Keypoint Transfer Experiment} \\
In addition to the real-robot experiments, we compare the performance of different functional keypoint transfer strategies, with a focus on the function point transfer. \\

\noindent \textbf{Baselines.} We evaluate four function point transfer strategies: 
\begin{itemize}
    \item Demo+VLM+DSC (proposed), which utilizes demonstration functional keypoints as references to prompt the VLM for region proposal, followed by point transfer through a dense semantic correspondence model;
    \item Demo+VLM, which removes the dense semantic correspondence model from the proposed implementation;
    \item Demo+DSC, which relies solely on a dense semantic correspondence model for functional keypoint transfer;
    \item VLM (zero-shot), which directly prompts the VLM to propose functional keypoints in a zero-shot manner. \\
\end{itemize}

\begin{figure}[h]
  \centering
  \begin{tikzpicture}[inner sep = 0pt, outer sep = 0pt]
    \node[anchor=south west] (fnC) at (0in,0in)
      {\includegraphics[height=1.3in,clip=true,trim=0in 0in 0in 0in]{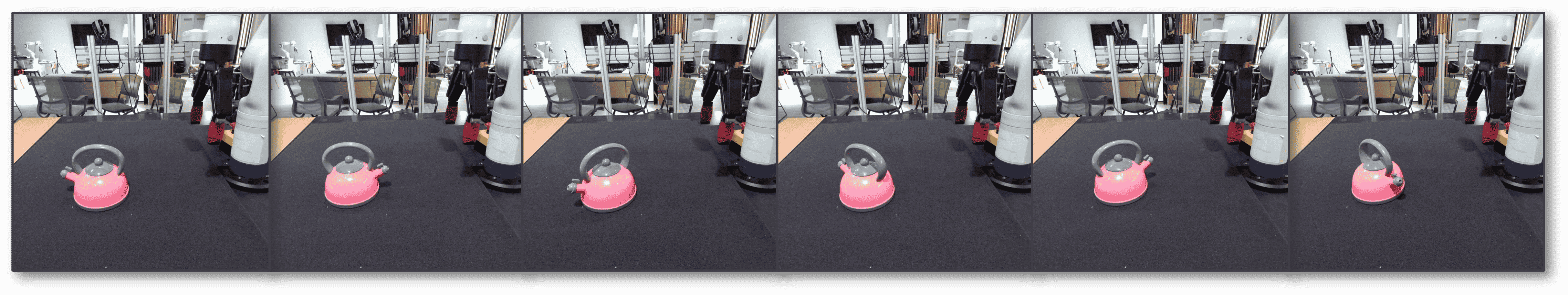}};
  \end{tikzpicture}
    \vspace*{-0.2in}
  \caption{Examples of collected images for function point transfer evaluation.}
  \label{fig:keypoint_view}
\end{figure}

\noindent \textbf{Experimental Setup.} For each test tool used in the real-robot experiment, we capture RGB images from 6 different views, covering various positions and orientations within the workspace. Each image has a resolution of 1280*720. A total of 150 images are used for evaluation. A set of examples is shown in Figure \ref{fig:keypoint_view}. 

\noindent \textbf{Evaluation Protocol.} To collect ground truth for function point transfer evaluation,  five volunteers were asked to annotate keypoints on test images using demonstration function points as references. Two evaluation metrics are used: (1) Average Keypoint Distance (AKD), which measures the average pixel distance between ground truth and detected keypoints. (2) Average Precision (AP), which represents the proportion of correctly detected keypoints under various thresholds. AP is evaluated under three thresholds: 15, 30, and 45 pixels. \\

\begin{table}[h]
\centering
\renewcommand\arraystretch{1.5}
\setlength\tabcolsep{3pt}
  \vspace*{-0.1in}
\caption{Quantitative results of Function Point Transfer}
\begin{tabular}{ccccc}
\toprule
\multirow{2}{*}{\textbf{Method}} & \multirow{2}{*}{AKD (pixel) $\downarrow$} & \multirow{2}{*}{AP@15 (\%) $\uparrow$} & \multirow{2}{*}{AP@30 (\%)$\uparrow$} & \multirow{2}{*}{AP@45 (\%)$\uparrow$} \\
                                 &                              &                            &                            &                            \\ \hline
Demo+VLM                         & 26.42                        & 38.89                      & 68.44                      & 83.56                      \\
Demo+DSC                         & 33.54                        & 47.11                      & 68.67                      & 78.67                      \\
VLM (zero-shot)                  & 56.09                        & 15.56                      & 36.22                      & 52.67                      \\
Demo+VLM+DSC (proposed)          & \textbf{18.54}                        & \textbf{51.33}                      & \textbf{85.78}                      & \textbf{94.44}                      \\ \bottomrule
\end{tabular}
\label{tab:func_transfer}
\end{table}

\noindent \textbf{Quantitative results.} The quantitative results of function point transfer are presented in Table \ref{tab:func_transfer}. The proposed Demo+VLM+DCS consistently outperforms the ablated strategies in both AKD and AP metrics. Demo+VLM achieves reasonable performance by leveraging the rich commonsense knowledge embedded in VLMs. However, VLMs alone struggle to provide precise point-level correspondences, which limits the effectiveness of Demo+VLM compared to the proposed strategy. Meanwhile, relying solely on the dense semantic correspondence model (i.e., Demo+DSC) often fails when faced with large intra-function variations. The performance gap between Demo+VLM and VLM (zero-shot) highlights the importance of demonstration functional keypoints, which serve as valuable references for proposing test functional keypoints.

\newpage

\noindent \textbf{Qualitative results.} 
\begin{figure}[h]
  \centering
  \begin{tikzpicture}[inner sep = 0pt, outer sep = 0pt]
    \node[anchor=south west] (fnC) at (0in,0in)
      {\includegraphics[height=5in,clip=true,trim=0in 0in 0in 0in]{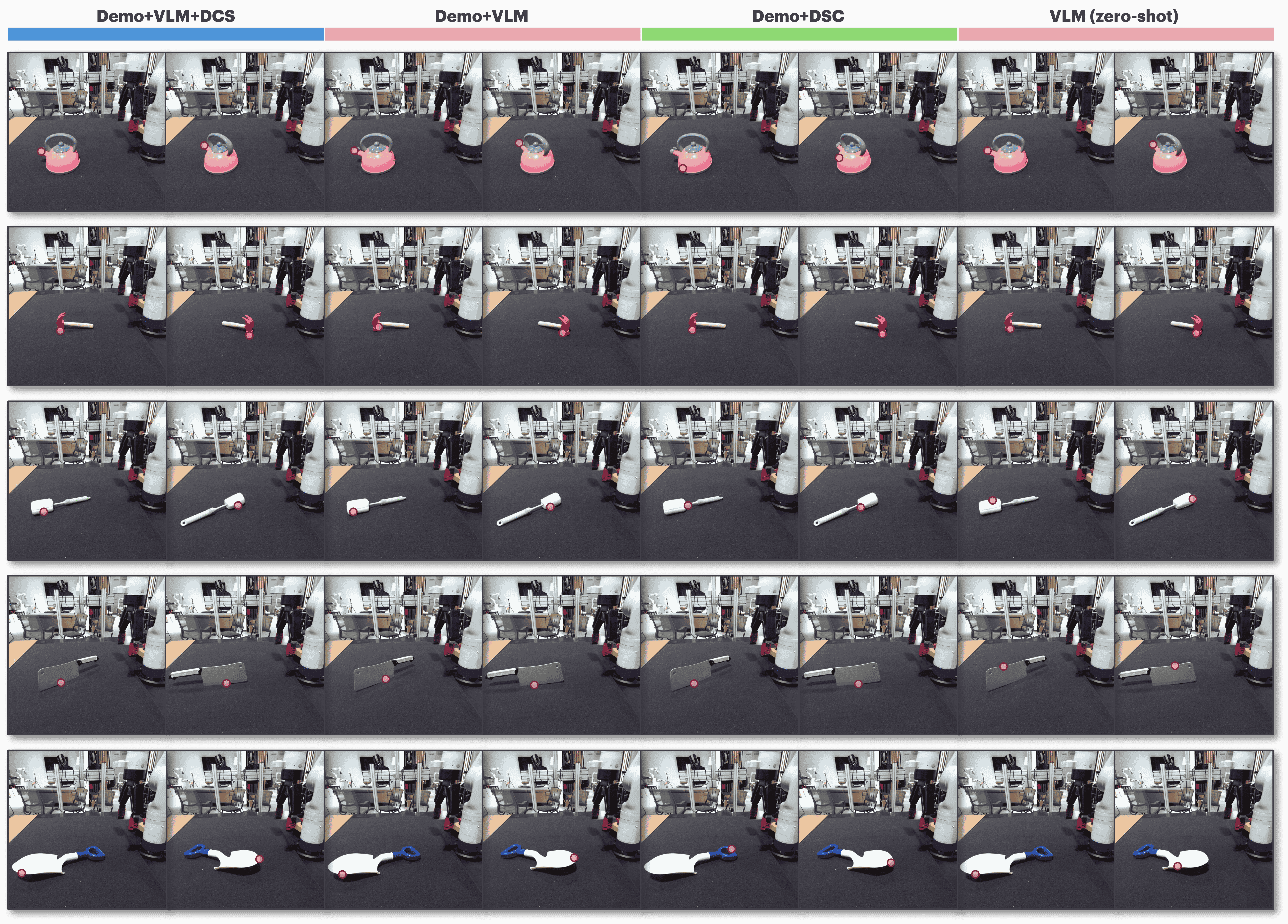}};
  \end{tikzpicture}
    \vspace*{-0.1in}
  \caption{Qualitative results of function point transfer.}
  \label{fig:keypoint_qualitative}
\end{figure}
\newpage


\noindent \textbf{D. Function-Centric Correspondence Implementation Detail}\\
This section provides the implementation details for functional keypoint transfer and function-centric correspondence. \\

\noindent \textbf{Function Keypoint Transfer.} The pseudo-code for functional keypoint transfer is illustrated in Algorithm \ref{alg:3d_keypoint_transfer}.

\begin{algorithm}[h]
\caption{Functional Keypoint Transfer.}
\label{alg:3d_keypoint_transfer}
\textbf{Input:} \\
\hspace{1em} Demo functional keypoints \( K_H^0 = [p_{\text{func}}^0, p_{\text{grasp}}^0, p_{\text{center}}^0] \), Initial keyframe \( I_0 \), Robot observation \( o_R \), Test tool mask \( M \), \\
\hspace{1em} Vision-language model (VLM), Dense semantic correspondence model \( \Phi \), \\
\hspace{1em} 3D-2D projection \( P_{\text{3D-2D}} \), 2D-3D projection \( P_{\text{2D-3D}} \), 3D center computation \( F_{\text{center}} \) \\
\textbf{Output:} Test functional keypoints \( K_R^0 = [q_{\text{func}}^0, q_{\text{grasp}}^0, q_{\text{center}}^0] \)

\begin{algorithmic}[1]
    \State \( K_R \gets \emptyset \)

    \State \textbf{1. Coarse-Grained Region Proposal:}
    \For{each \( k \in \{\text{func}, \text{grasp}\} \)}
        \State \( p_k^{2D} \gets P_{\text{3D-2D}}(p_k^0, I_0) \)
        \State \( r_k \gets \text{VLM}(p_k^{2D}, I_0, o_R, M) \) \Comment{Region proposal}
    \EndFor

    \State \textbf{2. Fine-Grained Point Transfer:}
    \For{each \( k \in \{\text{func}, \text{grasp}\} \)}
        \State \( q_k^{2D} \gets \Phi(p_k^{2D}, r_k, I_0, o_R) \) \Comment{Point transfer}
        \State \( q_k^0 \gets P_{\text{2D-3D}}(q_k^{2D}, o_R) \)
    \EndFor

    \State \textbf{3. 3D Center Computation:}
    \State \( q_{\text{center}}^0 \gets F_{\text{center}}(M, o_R) \)

    \State \textbf{4. Functional Keypoint Transfer Output:}
    \State \( K_R^0 \gets [q_{\text{func}}^0, q_{\text{grasp}}^0, q_{\text{center}}^0] \)
\end{algorithmic}
\end{algorithm}

\noindent \textbf{Function Plane Construction.} We aim to construct function planes $\Pi_H^{t_f}$  and  $\Pi_R^0$  based on the functional keypoints $ K_H^{t_f} = [p_{\text{func}}^{t_f}, p_{\text{grasp}}^{t_f}, p_{\text{center}}^{t_f}]$  and  $ K_R^0 = [q_{\text{func}}^0, q_{\text{grasp}}^0, q_{\text{center}}^0]$. $\Pi_H^{t_f}$ are defined by the following vectors:

\begin{enumerate}
    \item \textbf{Function Axis} 
    \begin{itemize}
        \item Definition: \[
    \mathbf{u}_H^{t_f} = \frac{p_{\text{func}}^{t_f} - p_{\text{center}}^{t_f}}{\|p_{\text{func}}^{t_f} - p_{\text{center}}^{t_f}\|}
    \]
    \item Description:  $\mathbf{u}_H^{t_f}$  is a normalized vector that defines the function axis. It points from the center point $p_{\text{center}}^{t_f}$ to the function point $p_{\text{func}}^{t_f}$ at $t_f$. This axis represents the primary direction along which the function operates.    
    \end{itemize}
    \item \textbf{Grasp Vector}
    \begin{itemize}
        \item Definition: \[  
    \mathbf{v}_H^{t_f} = \frac{p_{\text{grasp}}^{t_f} - p_{\text{func}}^{t_f}}{\|p_{\text{grasp}}^{t_f} - p_{\text{func}}^{t_f}\|}
        \]
    \item Description: $\mathbf{v}_H^{t_f}$  is a normalized vector that points from the function point $p_{\text{func}}^{t_f}$ to the grasp point $p_{\text{grasp}}^{t_f}$ at $t_f$.
    \end{itemize}
    \item \textbf{Normalized Normal Vector}
    \begin{itemize}
        \item Definition:
        \[ 
    \mathbf{n}_H^{t_f} = \frac{\mathbf{u}_H^{t_f} \times \mathbf{v}_H^{t_f}}{\| \mathbf{u}_H^{t_f} \times \mathbf{v}_H^{t_f} \|}
        \]
        \item Description: $\mathbf{n}_H^{t_f}$ is the unit normal vector of the function plane $\Pi_H^{t_f}$.
    \end{itemize}
    \item \textbf{Function Plane}
    \begin{itemize}
        \item Definition:
        \[   
    \Pi_H^{t_f}: (\mathbf{p} - p_{\text{func}}^{t_f}) \cdot \mathbf{n}_H^{t_f} = 0
        \]
        \item Description: The function plane is defined by the function point and its normal vector, describing the tool’s orientation and spatial configuration at $t_f$. 
    \end{itemize}
\end{enumerate}
Similarly, \(\mathbf{u}_R^0\), \(\mathbf{v}_R^0\), and \(\mathbf{n}_R^0\) are defined for \( \Pi_R^0 \).\\

\begin{figure}[h]
  \centering
  \begin{tikzpicture}[inner sep = 0pt, outer sep = 0pt]
    \node[anchor=south west] (fnC) at (0in,0in)
      {\includegraphics[height=1.1in,clip=true,trim=0.2in 0in 0in 0in]{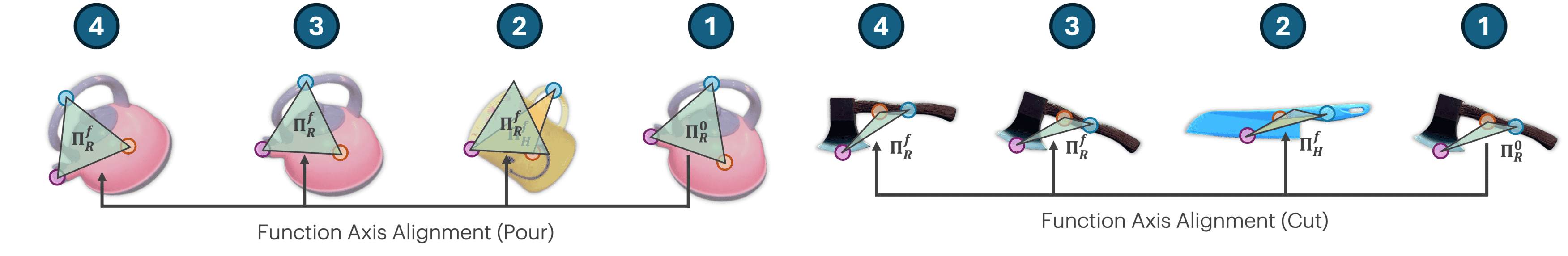}};
  \end{tikzpicture}
    \vspace*{-0.1in}
  \caption{Examples of function axis alignment.}
  \label{fig:axis_example}
\end{figure}

\noindent \textbf{Function Axis Alignment.} Simply aligning the function axes of the demonstration and test tools may not yield a feasible function keyframe pose for the test tool. This is due to substantially different relative locations of the three functional keypoints (particularly for cross-category generalization). As is shown in Figure \ref{fig:axis_example}, the two function keyframe poses in Step 3 may fail to achieve successful task executions, as the teapot and axe are not tilted sufficiently.

\begin{figure}[h]
  \centering
  \begin{tikzpicture}[inner sep = 0pt, outer sep = 0pt]
    \node[anchor=south west] (fnC) at (0in,0in)
      {\includegraphics[height=0.7in,clip=true,trim=0in 0in 0in 0in]{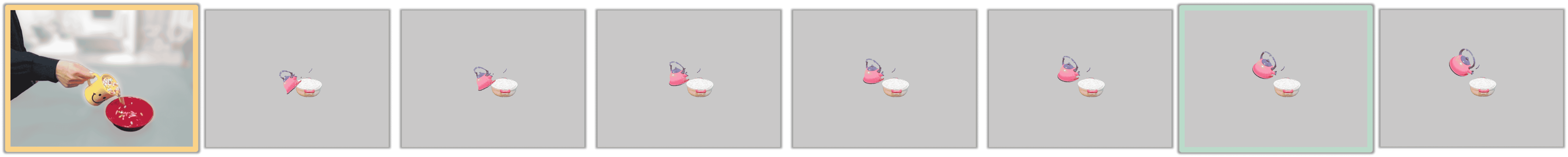}};
  \end{tikzpicture}
  \caption{Examples of rendered synthetic function keyframes. The demonstration function keyframe is highlighted in yellow, and the selected test function keyframe is highlighted in green.}
  \label{fig:axis_render}
\end{figure}

\noindent To address this issue, we refine the function axis alignment using the VLM. Specifically, in Step 3, we rotate the function plane with the function point as the origin and the normal vector as the rotation axis. Seven angle offsets ranging from $[-45^{\circ}, -45^{\circ}]$ are applied , including $-45^{\circ}$, $-30^{\circ}$, $-10^{\circ}$, $0^{\circ}$, $10^{\circ}$, $30^{\circ}$, $45^{\circ}$. Next, the combined point cloud of each rotated test tool and the target are back-projected onto the camera planes using the camera intrinsic matrix, rendering seven synthetic 
function keyframes. Examples of rendered synthetic function keyframes are presented in Figure \ref{fig:axis_render}. Then, we prompt the VLM, using the demonstration function keyframe as the reference, to  identify the image that represents the optimal state conducive to the task success. The detailed prompt is given in Appendix G. Finally, the rotation transformation corresponding to the optimal function keyframe state is recorded for function axis alignment (Step 4). 

\newpage




\noindent \textbf{E. Trajectory Optimization Implementation Detail} \\
In the section, we provide implementation details for trajectory optimization, complementing the constrained optimization problem  formulated in the manuscript.  \\

\begin{figure}[h]
  \centering
  \begin{tikzpicture}[inner sep = 0pt, outer sep = 0pt]
    \node[anchor=south west] (fnC) at (0in,0in)
      {\includegraphics[height=2.5in,clip=true,trim=0in 0in 0in 0in]{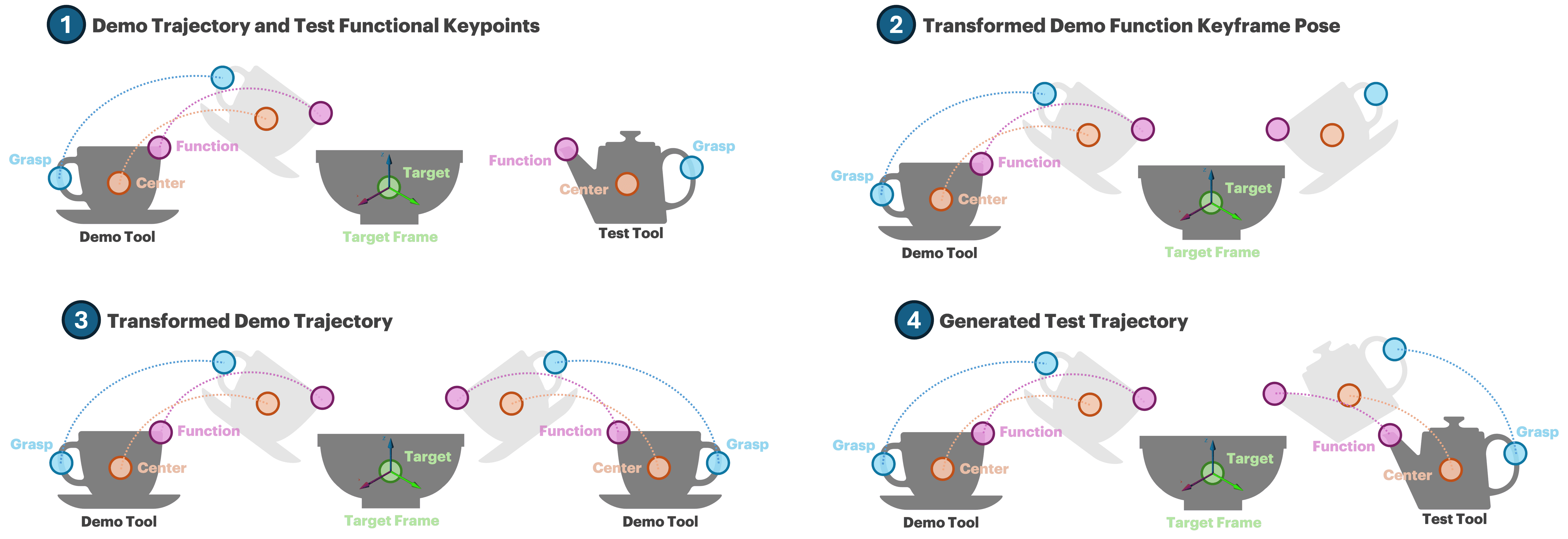}};
  \end{tikzpicture}
    \vspace*{-0.1in}
  \caption{Demonstration trajectory and function keyframe pose wrapping.}
  \label{fig:init_align}
\end{figure}

\noindent \textbf{Demonstration Trajectory and Pose Wrapping.} As shown in Figure \ref{fig:init_align}, Step 1, the demonstration and test tools are positioned on opposite sides of the target. While the demonstration trajectory requires the test tool to approach from the left side for pouring, the target object’s rotational symmetry about the z-axis allows approaching from the right side as well. This symmetry, common among target objects in the experiment, enables us to warp the demonstration trajectory and function keyframe pose to generate shorter and easier-to-execute test tool trajectories.

\noindent In Figure \ref{fig:init_align}, Step 2, we align the demonstration function frame pose with the test tool by rotating the demonstration functional keypoints around the z-axis. The rotation angle is computed based on the angular difference between the demonstration and test function points. Similarly, in Step 3, we wrap the demonstration trajectory through two operations:  (1) aligning it with the test tool by rotating around the z-axis and (2) scaling its translational component based on the test tool’s function point. Finally, the wrapped demonstration trajectory and function keyframe pose serve as the reference and constraint for trajectory optimization (Step 4).  \\

\noindent \textbf{Optimization Constraints and Costs.} In addition to the trajectory cost and the keyframe pose constraints described in the manuscript, we incorporate the following adjustments: 

\begin{itemize}
    \item Early Trajectory Cost Relaxation. The trajectory cost is omitted for the initial 30\% of the trajectory, as the interaction primarily occurs during the later phases. This approach also allows the optimizer to explore more feasible paths and ensure smoother transitions to the interaction phase, particularly when the initial states of the demonstration and test tools differ significantly.
    \item Velocity Constraint. We impose limits on the translational and angular velocities of the test tool to ensure smooth and physically feasible trajectory generation.
    \item Collision Avoidance Constraint. This constraint enforces a minimum Euclidean distance between the test tool and the 3D bounding box of the obstacle to prevent collisions during trajectory execution. 
\end{itemize}
\noindent We employ CasADi as the optimization framework for symbolic modeling and automatic differentiation. IPOPT is used as the solver to efficiently handle the nonlinear programming problem with constraints.
\newpage

\noindent \textbf{F. PD Controller Implementation Detail} \\
This section details the practical implementation, control architecture, and parameter selection for the implemented velocity-based PD controller.\\

\noindent \textbf{Controller Architecture.} The controller implements joint velocity control with null space optimization.
The input commands are received at $10\text{ Hz}$ while the controller operates at a higher frequency of $200\text{ Hz}$. To ensure smooth motion, we implement trajectory interpolation between commanded poses. \\

\noindent The PD control law is formulated separately for translation and rotation. 

\begin{enumerate}
    \item For the translational motion: 
    \[
        \mathbf{v}_{\text{lin}} = \begin{bmatrix}
            v_x \\
            v_y \\
            v_z
        \end{bmatrix} = \underbrace{\mathbf{K}_p}_{\text{proportional}} \underbrace{\begin{bmatrix}
            e_{p,x} \\
            e_{p,y} \\
            e_{p,z}
        \end{bmatrix}}_{\text{position error}} + \underbrace{\mathbf{K}_d}_{\text{derivative}} \underbrace{\begin{bmatrix}
            \dot{e}_{p,x} \\
            \dot{e}_{p,y} \\
            \dot{e}_{p,z}
        \end{bmatrix}}_{\text{velocity error}}
    \]
    \item For the rotational motion:
    \[
        \boldsymbol{\omega} = \begin{bmatrix}
            \omega_x \\
            \omega_y \\
            \omega_z
        \end{bmatrix} = \underbrace{\mathbf{K}_{p,\text{rot}}}_{\text{proportional}} \underbrace{\begin{bmatrix}
            e_{\theta,x} \\
            e_{\theta,y} \\
            e_{\theta,z}
        \end{bmatrix}}_{\text{orientation error}} + \underbrace{\mathbf{K}_{d,\text{rot}}}_{\text{derivative}} \underbrace{\begin{bmatrix}
            \dot{e}_{\theta,x} \\
            \dot{e}_{\theta,y} \\
            \dot{e}_{\theta,z}
        \end{bmatrix}}_{\text{angular velocity error}}
    \]
\end{enumerate}

\noindent where the control gains are represented by the following  matrices:
\[
    \mathbf{K}_p = \begin{bmatrix} 
        3.0 & 0 & 0 \\
        0 & 3.0 & 0 \\
        0 & 0 & 3.0
    \end{bmatrix} \in \mathbb{R}^{3 \times 3}, \quad
    \mathbf{K}_d = \begin{bmatrix}
        0.001 & 0 & 0 \\
        0 & 0.001 & 0 \\
        0 & 0 & 0.001
    \end{bmatrix} \in \mathbb{R}^{3 \times 3}
\]
\[
    \mathbf{K}_{p,\text{rot}} = \begin{bmatrix}
        3.0 & 0 & 0 \\
        0 & 3.0 & 0 \\
        0 & 0 & 3.0
    \end{bmatrix} \in \mathbb{R}^{3 \times 3}, \quad
    \mathbf{K}_{d,\text{rot}} = \begin{bmatrix}
        0.01 & 0 & 0 \\
        0 & 0.01 & 0 \\
        0 & 0 & 0.01
    \end{bmatrix} \in \mathbb{R}^{3 \times 3}
\]
\\
\noindent \textbf{Trajectory Interpolation.} To address the frequency mismatch between command inputs and controller execution, we implement a trajectory interpolation scheme. Given two consecutive desired poses at times $t_k$ and $t_{k+1}$:
\[
    \mathbf{X}_k = \begin{bmatrix} \mathbf{p}_k \\ \mathbf{q}_k \end{bmatrix}, \quad \mathbf{X}_{k+1} = \begin{bmatrix} \mathbf{p}_{k+1} \\ \mathbf{q}_{k+1} \end{bmatrix}
\]

\noindent where $\mathbf{p}$ represents position and $\mathbf{q}$ represents orientation in quaternion form. 

\noindent The number of interpolation points is determined by:
\[
    N = \max(1, \lfloor (t_{k+1} - t_k) f_c \rfloor)
\]

\noindent For position interpolation, we use linear interpolation:
\[
    \mathbf{p}(s) = (1-s)\mathbf{p}_k + s\mathbf{p}_{k+1}, \quad s \in [0,1]
\]

\noindent For orientation interpolation, we use spherical linear interpolation (SLERP):
\[
    \mathbf{q}(s) = \frac{\sin((1-s)\Omega)}{\sin(\Omega)}\mathbf{q}_k + \frac{\sin(s\Omega)}{\sin(\Omega)}\mathbf{q}_{k+1}
\]

\noindent where $\Omega = \arccos(\mathbf{q}_k \cdot \mathbf{q}_{k+1})$ is the angle between quaternions. 

\noindent The interpolation parameter $s$ is discretized as:
\[
    s_i = \frac{i}{N-1}, \quad i = 0,\ldots,N-1
\]
\\
\noindent \textbf{Error Computation.} The translational error is computed directly in Cartesian space:
\[
    \mathbf{e}_p = \mathbf{x}_{\text{des}} - \mathbf{x}_{\text{cur}}
\]

\noindent For safety, the controller implements a position error threshold:
\[
    \|\mathbf{e}_p\| \leq 0.5 \text{ m}
\]

\noindent The rotational error is computed using rotation matrices:
\[
    \mathbf{R}_{\text{error}} = \mathbf{R}_{\text{des}} \mathbf{R}_{\text{cur}}^{-1}
\]

\noindent The error is converted to axis-angle representation and normalized to ensure the rotation angle remains within $[-\pi, \pi]$:
\[
    \mathbf{e}_\theta = \begin{cases}
        \mathbf{e}_{\text{axis-angle}} & \text{if } \|\mathbf{e}_{\text{axis-angle}}\| \leq \pi \\
        \mathbf{e}_{\text{axis-angle}} \frac{\|\mathbf{e}_{\text{axis-angle}}\| - 2\pi}{\|\mathbf{e}_{\text{axis-angle}}\|} & \text{otherwise}
    \end{cases}
\]
\\
\noindent \textbf{Joint Space Control.} For the joint velocity control mode, the Cartesian velocities are mapped to joint space using the manipulator Jacobian:
\[
    \dot{\mathbf{q}} = \mathbf{J}^\dagger \begin{bmatrix} \mathbf{v}_{\text{lin}} \\ \boldsymbol{\omega} \end{bmatrix} + \mathbf{N}\dot{\mathbf{q}}_0
\]

\noindent where:
\begin{itemize}
    \item $\mathbf{J}^\dagger$ is the Moore-Penrose pseudoinverse of the Jacobian
    \item $\mathbf{N} = \mathbf{I} - \mathbf{J}^\dagger\mathbf{J}$ is the null space projector
    \item $\dot{\mathbf{q}}_0$ is the null space velocity\\
\end{itemize}

\noindent \textbf{Null Space Optimization.} The null space velocity combines two objectives:
\[
    \dot{\mathbf{q}}_0 = K_{\text{home}}\left(\mathbf{q}_{\text{home}} - \mathbf{q}_{\text{cur}}\right) - K_{\text{min}}\mathbf{q}_{\text{cur}}
\]

\noindent where:
\begin{itemize}
    \item $K_{\text{home}} = 0.1$ is the gain for home configuration attraction
    \item $K_{\text{min}} = 0.05$ is the gain for joint velocity minimization
    \item $\mathbf{q}_{\text{home}}$ is the preferred home configuration
\end{itemize}

\newpage

\noindent \textbf{G. VLM Prompting Implementation Detail}

\begin{center}
\begin{tcolorbox}[colback=gray!5, colframe=black!40, sharp corners=south, title= Function Point Detection Prompt]\small

Given an interaction frame between two objects, select pre-defined keypoints. \\

The input request contains:  
\begin{itemize}
    \item The task information as dictionaries. The dictionary contains these fields: 
    \begin{itemize}
        \item `\textbf{instruction}': The task in natural language forms.
        \item `\textbf{object\_grasped}': The object that the human holds in hand while executing the task.
        \item `\textbf{object\_unattached}': The object that the human will interact with `object\_grasped' without holding it in hand.
    \end{itemize}
    \item  An image of the current table-top environment captured from a third-person view camera, annotated with a set of visual marks:
    \begin{itemize}
        \item \textbf{candidate keypoints on `object\_grasped'}: Red dots marked as `P\textsubscript{i}' on the image, where [i] is an integer. \\
    \end{itemize}
\end{itemize}

The interaction is specified by `function\_keypoint' on the `object\_grasped':
\begin{itemize}
    \item The human hand grasps `object\_grasped' and moves the `function\_keypoint' to approach `object\_unattached'.
    \item `\textbf{function\_keypoint}': The point on `object\_grasped' indicating the part that will contact `object\_unattached'. \\
\end{itemize}

The response should be a dictionary in JSON form, which contains:
\begin{itemize}
    \item `\textbf{function\_keypoint}': Selected from candidate keypoints marked as `P\textsubscript{i}' on the image.\\
\end{itemize}

Think about this step by step:
\begin{enumerate}[leftmargin=*, label=\arabic*.]
    \item Describe the region where interaction between `object\_grasped' and `object\_unattached' happens.
    \item Select `function\_keypoint' on the `object\_grasped' within the interaction region.
\end{enumerate}
\end{tcolorbox}
\end{center}

\begin{center}
\begin{tcolorbox}[colback=gray!5, colframe=black!40, sharp corners=south, title= Function Point Transfer Prompt]\small

Refer to the position of red keypoint on the first example image, select corresponding pre-defined keypoints on the second test image. \\

The input request contains:  
\begin{itemize}
    \item The task information as dictionaries. The dictionary contains these fields: 
    \begin{itemize}
        \item `\textbf{instruction}': The task in natural language forms.
        \item `\textbf{object\_grasped}': The object that the human holds in hand while executing the task.
        \item `\textbf{object\_unattached}': The object that the human will interact with `object\_grasped' without holding it in hand.
    \end{itemize}
    \item An example image annotated with a red keypoint.
    \item A test image of the current table-top environment captured from a third-person view camera, annotated with a set of visual marks:
    \begin{itemize}
        \item \textbf{candidate keypoints on `object\_grasped'}: Red dots marked as `P\textsubscript{i}' on the image, where [i] is an integer. \\
    \end{itemize}
\end{itemize}

The interaction is specified by `function\_keypoint' on the `object\_grasped':
\begin{itemize}
    \item Select the candidate keypoint on the test image corresponds to the red keypoint annotated on the example image.
    \item `\textbf{function\_keypoint}': The point on `object\_grasped' indicating the part that will contact `object\_unattached'. \\
\end{itemize}

The response should be a dictionary in JSON form, which contains:
\begin{itemize}
    \item `\textbf{function\_keypoint}': Selected from candidate keypoints marked as `P\textsubscript{i}' on the image. \\
\end{itemize}

Think about this step by step:
\begin{enumerate}[leftmargin=*, label=\arabic*.]
    \item Describe the object part where keypoint is located on the example image.
    \item Describe the region where interaction between `object\_grasped' and `object\_unattached' happens.
    \item Select `function\_keypoint' on the `object\_grasped' within the interaction region on the test image.
\end{enumerate}
\end{tcolorbox}
\end{center}

\newpage

\begin{center}
\begin{tcolorbox}[colback=gray!5, colframe=black!40, sharp corners=south, title= Grasp Point Transfer Prompt]\small

Refer to the position of red keypoint on the first example image, select corresponding pre-defined keypoints on the second test image. \\

The input request contains:  
\begin{itemize}
    \item The task information as dictionaries. The dictionary contains these fields: 
    \begin{itemize}
        \item `\textbf{instruction}': The task in natural language forms.
        \item `\textbf{object\_grasped}': The object that the human holds in hand while executing the task.
        \item `\textbf{object\_unattached}': The object that the human will interact with `object\_grasped' without holding it in hand.
    \end{itemize}
    \item An example image annotated with a red keypoint.
    \item A test image of the current table-top environment captured from a third-person view camera, annotated with a set of visual marks:
    \begin{itemize}
        \item \textbf{candidate keypoints on `object\_grasped'}: Red dots marked as `P\textsubscript{i}' on the image, where [i] is an integer. \\
    \end{itemize}
\end{itemize}

The interaction is specified by `grasp\_keypoint' on the `object\_grasped':
\begin{itemize}
    \item Select the candidate keypoint on the test image corresponds to the red keypoint annotated on the example image.
    \item The human hand grasps the `object\_grasped' at the `grasp\_keypoint’.
    \item `\textbf{grasp\_keypoint}': The point on `object\_grasped' indicates the part where the hand should hold. \\
\end{itemize}

The response should be a dictionary in JSON form, which contains:
\begin{itemize}
    \item `\textbf{grasp\_keypoint}': Selected from candidate keypoints marked as `P\textsubscript{i}' on the image. \\
\end{itemize}

Think about this step by step:
\begin{enumerate}[leftmargin=*, label=\arabic*.]
    \item Describe the object part where keypoint is located on the example image.
    \item Find the part on `object\_grasped’ where humans usually grasp.
    \item Select `grasp\_keypoint' on the `object\_grasped' within the interaction region on the test image.
\end{enumerate}
\end{tcolorbox}
\end{center}

\begin{center}
\begin{tcolorbox}[colback=gray!5, colframe=black!40, sharp corners=south, title= Function Axis Alignment Prompt]

From a list of interaction frames between the tool and target objects, select the image that represents the state most conducive to completing the task. \\

The input request contains:  
\begin{itemize}
    \item The task information as dictionaries. The dictionary contains these fields: 
    \begin{itemize}
        \item `\textbf{instruction}': The task in natural language forms.
        \item `\textbf{object\_grasped}': The object that the human holds in hand while executing the task.
        \item `\textbf{object\_unattached}': The object that the human will interact with `object\_grasped' without holding it in hand.
    \end{itemize}
    \item A list of interaction frames between the tool and target objects. \\
\end{itemize} 

The response should be a dictionary in JSON form, which contains:
\begin{itemize}
    \item `\textbf{selected\_idx}': the idx of the selected image.
\end{itemize}
\end{tcolorbox}
\end{center}

\newpage
\noindent \textbf{H. Q\&A Section} \\
In this section, we address common questions about our method, clarifying potential concerns, discussing limitations, and providing insights into its design, implementation, future improvements, and broader applicability. \\

\begin{itemize}
    \item \textbf{Q1: Can this method be applied to a wider range of tool manipulation tasks?} \\
    \textbf{A1:} FUNCTO is generally applicable to tool manipulation tasks involving two-object interactions (tool and target), where object dynamics are less critical to task success. Examples include peeling, sweeping, stirring, picking, placing, mixing, inserting, stacking, and flipping.\\

    \item \textbf{Q2: Can the proposed functional keypoint transfer strategy be extended to a broader range of applications beyond those demonstrated in the paper} \\
    \textbf{A2:} Yes, the proposed functional keypoint transfer strategy can be applied to other tasks involving semantic correspondences. Notably, we have adapted this approach for semantic keypoint transfer in cloth manipulation. \\

    \item \textbf{Q3: How can this method leverage additional demonstrations for improved performance?} \\
    \textbf{A3:} As discussed in the Limitations subsection, a function is inherently multi-modal. Therefore, FUNCTO can leverage few-shot human demonstrations for multi-modal modeling. During inference, the robot can retrieve the ``best" demonstration from the database to enhance task execution. \\

    \item \textbf{Q4: What are the benefits of explicitly representing skill-use keypoints and trajectories?} \\
    \textbf{A4:} The interpretable explicit representation can be integrated with existing task and motion planning algorithms, enabling the execution of long-horizon tool manipulation tasks. \\

    \item \textbf{Q5: What is the role of foundation models in this approach, and why are they essential?} \\
    \textbf{A5:} Foundation models provide commonsense knowledge, enabling the inference of information that cannot be directly extracted from geometric or visual cues. For instance, tools with the same function may exhibit significant intra-function variations. Transferring functional keypoints based solely on geometric or visual similarities prone to failures. Additionally, without the commonsense reasoning embedded in foundation models, the robot may struggle to accurately infer the correct functional axis alignment transformation. \\
\end{itemize}

\end{document}